\documentclass[10pt,twocolumn,letterpaper]{article}
\usepackage[pagenumbers]{iccv}
\usepackage[accsupp]{axessibility}
\usepackage{array}
\usepackage{arydshln}
\usepackage{bm}
\usepackage{boldline}
\usepackage{color}
\usepackage{dsfont}
\usepackage{enumitem}
\usepackage{float}
\usepackage{makecell}
\usepackage{mathtools}
\usepackage{multirow}
\usepackage{nicefrac}
\usepackage{pifont}
\usepackage[accsupp]{axessibility}

\definecolor{Urlcolor}{RGB}{251,111,146}
\definecolor{Linkcolor}{RGB}{193,18,31}
\definecolor{CiteColor}{RGB}{32,126,190}
\definecolor{darkgreen}{RGB}{0.0, 0.5, 0.0}
\usepackage[pagebackref,breaklinks,colorlinks,urlcolor=Urlcolor,linkcolor=Linkcolor,citecolor=CiteColor]{hyperref}

\makeatletter
\renewcommand{\paragraph}{%
    \@startsection{paragraph}{4}%
    {\z@}{-0.15em}{-0.5em}%
    {\normalfont\normalsize\bfseries}%
}
\makeatother

\newcommand{\method}{DSO\xspace}

\newcommand{\tablestyle}[2]{\setlength{\tabcolsep}{#1}\renewcommand{\arraystretch}{#2}\centering\footnotesize}

\def\paperID{2}
\def\confName{ICCV}
\def\confYear{2025}

\newcommand\rurl[1]{%
  \href{https://#1}{\nolinkurl{#1}}%
}

\title{\method: Aligning 3D Generators with Simulation Feedback for Physical Soundness}

\author{Ruining Li
\quad
Chuanxia Zheng
\quad
Christian Rupprecht
\quad
Andrea Vedaldi \\
Visual Geometry Group, University of Oxford \\
{\tt\small \{ruining, cxzheng, chrisr, vedaldi\}@robots.ox.ac.uk}\\[0.1em]
\small\rurl{ruiningli.com/dso}
}
\togglefalse{iccvpagenumbers}

\begin{document}
\twocolumn[\maketitle\vspace{-3em}{
\setlength{\fboxsep}{0pt}%
\setlength{\fboxrule}{0.1pt}%
\begin{center}
\fbox{\includegraphics[trim=0 0 0 0, clip, width=0.99\linewidth]{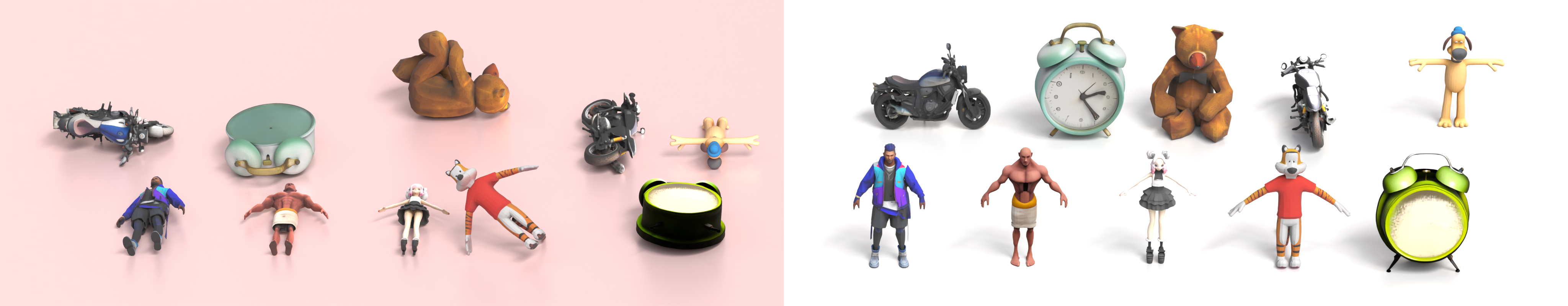}}
{
\newcolumntype{C}[1]{>{\centering}p{#1}}
\setlength\tabcolsep{0pt}
\begin{tabular}{C{.5\textwidth}C{.5\textwidth}}
Image-to-3D (TRELLIS) & Image-to-3D with \method (ours)
\end{tabular}
}
\end{center}
\vspace{-0.5em}
\begin{tabular}{cc}
\fbox{\includegraphics[height=5em,clip,trim=65cm 1cm 2cm 15cm]{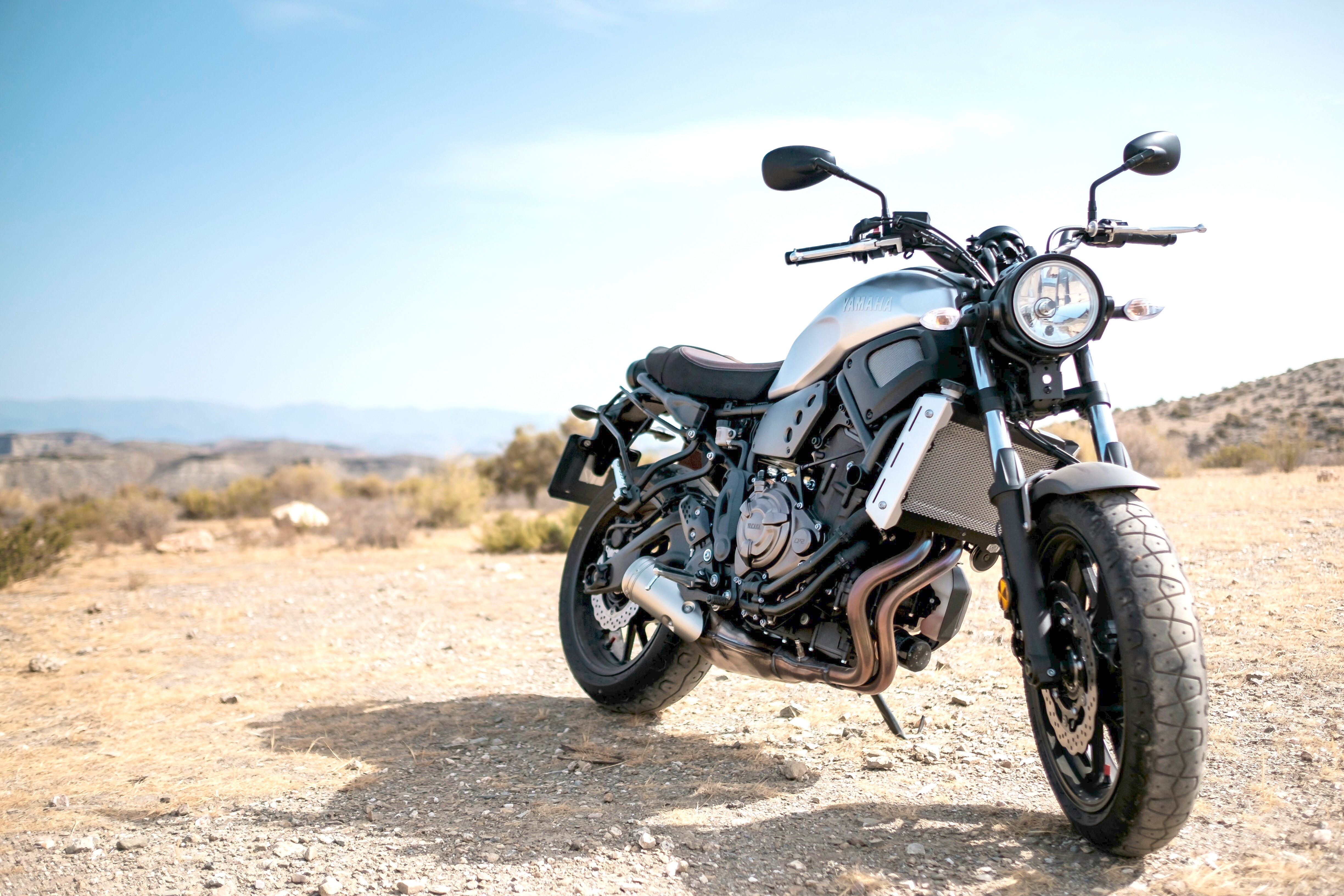}}~%
\fbox{\includegraphics[height=5em,clip,trim=67cm 12cm 67cm 8cm]{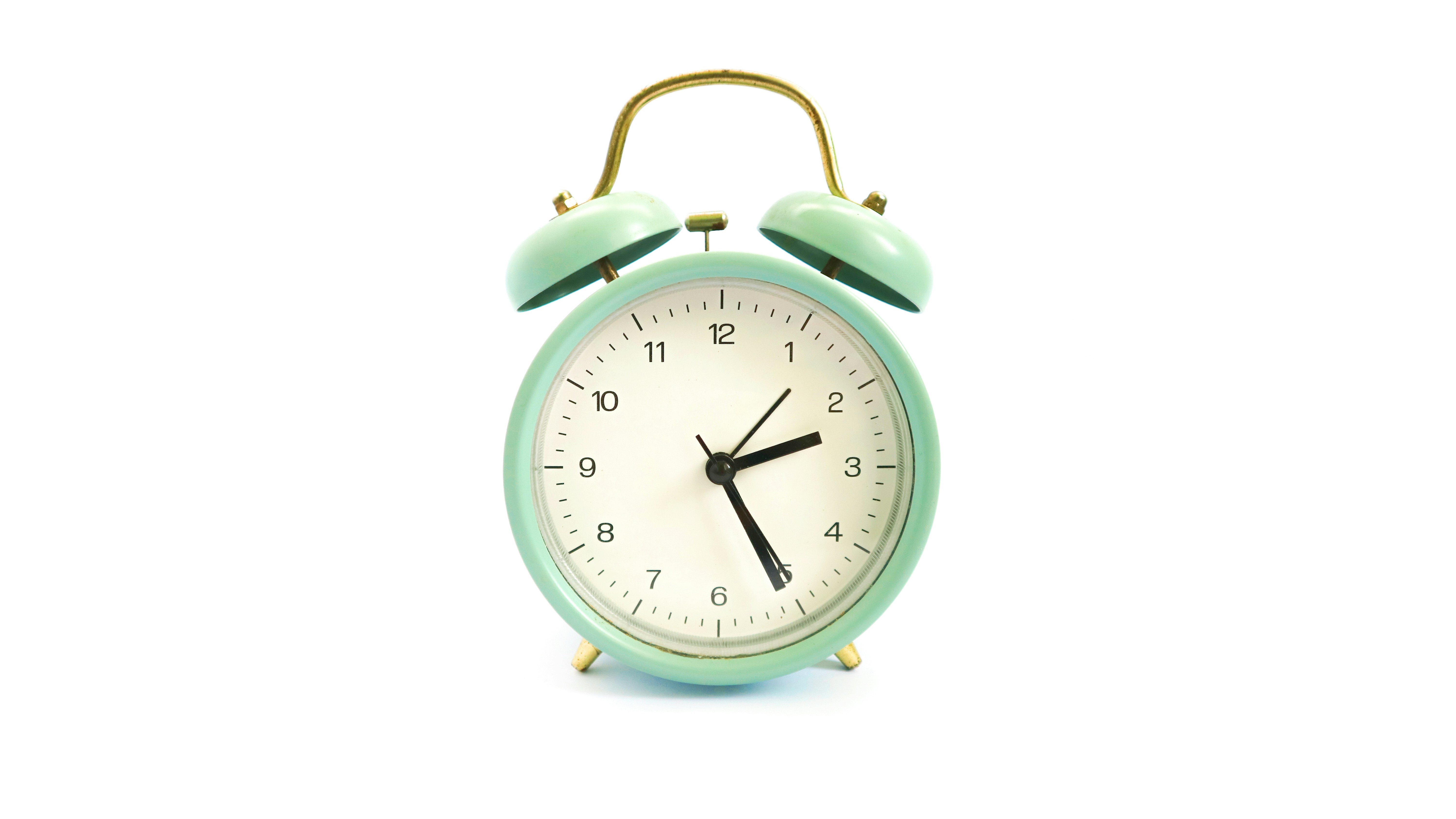}}~%
\fbox{\includegraphics[height=5em,clip,trim=2cm 1cm 2cm 1cm]{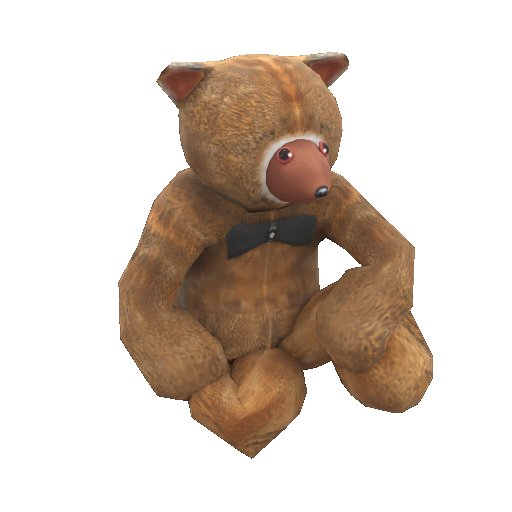}}~%
\fbox{\includegraphics[height=5em,clip,trim=0pt 30cm 0pt 50cm]{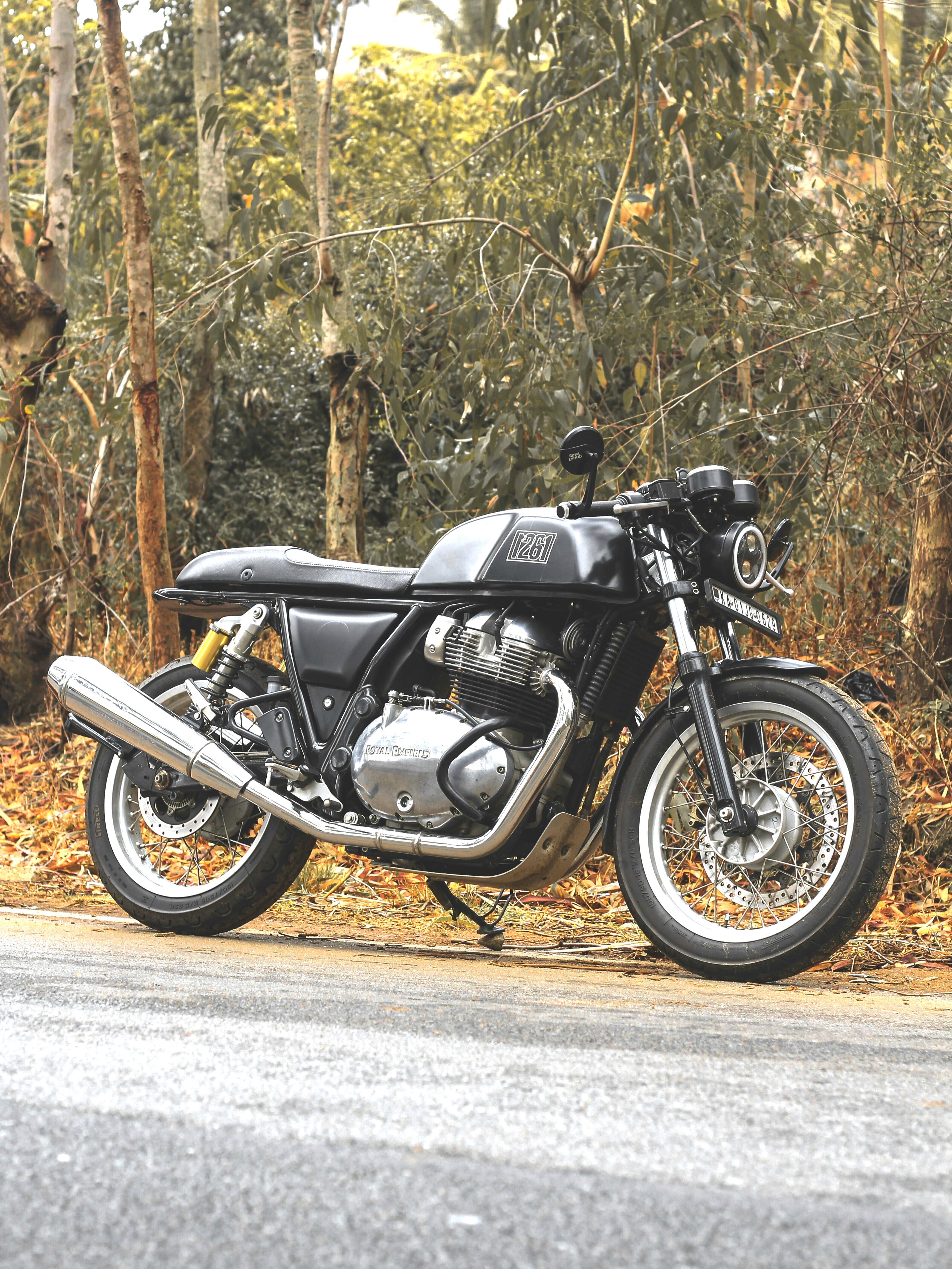}}~%
\fbox{\includegraphics[height=5em,clip,trim=3.5cm 3cm 1.5cm 2cm]{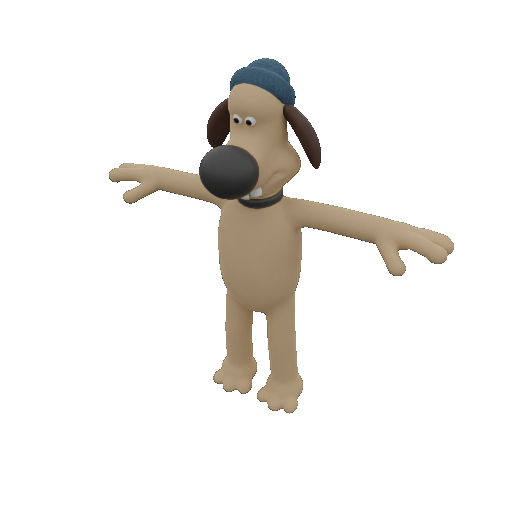}}%
&
\includegraphics[height=5em,clip,trim=0 2cm 0 3cm]{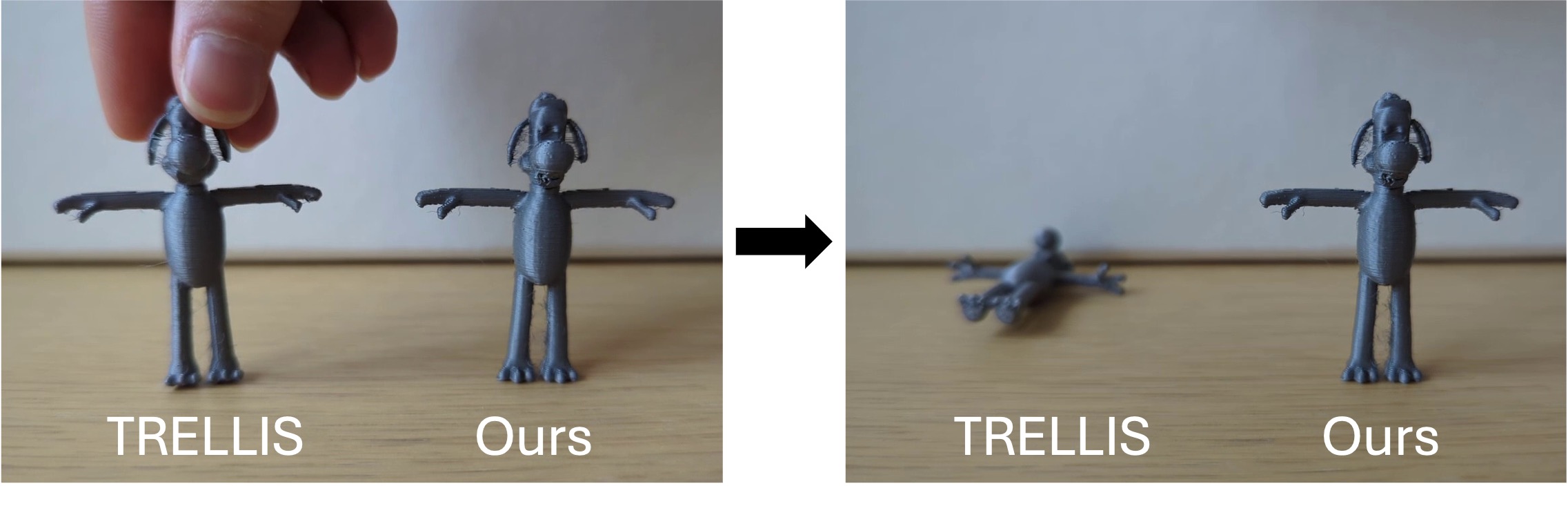}
\\
Input images & Real 3D prints
\end{tabular}
\captionof{figure}{
\emph{Top-left}: A state-of-the-art image-to-3D model like TRELLIS often fails to reconstruct 3D objects that can stand under gravity even when prompted with images of stable objects (\eg, \emph{bottom-left}).
\emph{Top-right}: Our method, \method, improves the image-to-3D model via \textbf{D}irect \textbf{S}imulation \textbf{O}ptimization, significantly increasing the likelihood that generated 3D objects can stand, in physical simulation and in real-life, when 3D printed (\emph{bottom-right}).
The method incurs no additional cost at test time, and can thus generate such objects in seconds.
}%
\label{fig:teaser}
}\bigbreak]
\begin{abstract}
Most 3D object generators prioritize aesthetic quality, often neglecting the physical constraints necessary for practical applications.
One such constraint is that a 3D object should be self-supporting, \ie, remain balanced under gravity.
Previous approaches to generating stable 3D objects relied on differentiable physics simulators to optimize geometry at test time, which is slow, unstable, and prone to local optima.
Inspired by the literature on aligning generative models with external feedback, we propose \textbf{D}irect \textbf{S}imulation \textbf{O}ptimization (DSO). This framework leverages feedback from a (non-differentiable) simulator to increase the likelihood that the 3D generator directly outputs stable 3D objects.
We construct a dataset of 3D objects labeled with stability scores obtained from the physics simulator.
This dataset enables fine-tuning of the 3D generator using the stability score as an alignment metric, via direct preference optimization (DPO) or direct reward optimization (DRO)—a novel objective we introduce to align diffusion models without requiring pairwise preferences.
Our experiments demonstrate that the fine-tuned \emph{feed-forward} generator, using either the DPO or DRO objective, is significantly faster and more likely to produce stable objects than test-time optimization.
Notably, the DSO framework functions even \emph{without} any ground-truth 3D objects for training, allowing the 3D generator to self-improve by automatically collecting simulation feedback on its own outputs.
\end{abstract}
\section{Introduction}%
\label{sec:intro}

\begin{figure}[tb!]
\centering
\includegraphics[width=\columnwidth, clip, trim=0 5 0 0]{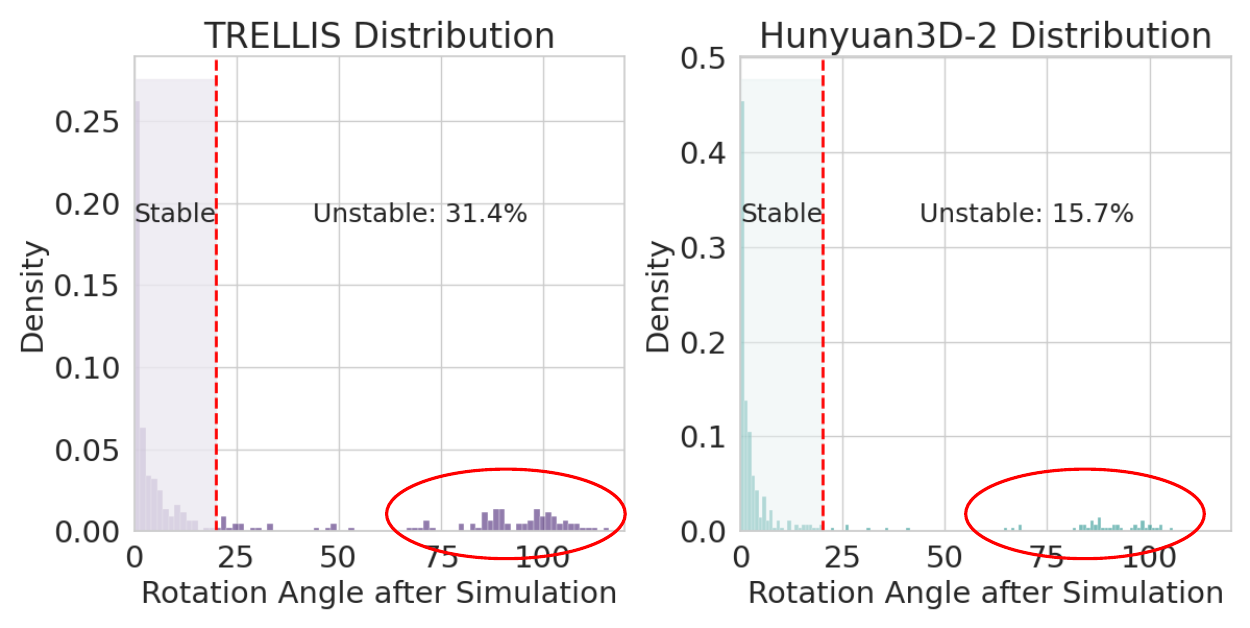}
\caption{
State-of-the-art 3D generators \emph{cannot} robustly produce stable objects.
Even when taking images of stable objects in their \emph{training} set as input, TRELLIS~\cite{xiang2024trellis} and Hunyuan3D-2.0~\cite{hunyuan3d22025tencent} generate about $30\%$ and $15\%$ unstable assets respectively.
}%
\label{fig:motivation}
\end{figure}

Given a single image of an object that is \emph{stable under gravity}, we consider the problem of reconstructing it in 3D.
Recent image-to-3D reconstructors~\cite{watsonnovel,liu2023zero,shi23mvdream,zheng2024free3d,voleti2025sv3d,liu2023one,zhang2024clay,xiang2024trellis,pooledreamfusion,lin2023magic3d,melas2023realfusion,yang2024hunyuan3d,hunyuan3d22025tencent,tang2025lgm,hong2023lrm} have focused on improving the quality of objects' 3D geometry and appearance, but not necessarily their physical soundness.
As shown in~\cref{fig:motivation}, when prompted with an image of a stable object, state-of-the-art generators like TRELLIS~\cite{xiang2024trellis} and Hunyuan3D 2.0~\cite{hunyuan3d22025tencent} often fail to produce a stable object in 3D.
The failure rate is 15\% even for objects seen during \emph{training} and increases significantly for new objects, such as the clock and motorcycles in~\cref{fig:teaser}.

Stability is a common property of natural and man-made objects and is important in many applications, such as fabrication and simulation~\cite{li2023behavior, Nasiriany2024RoboCasaLS}.
It is, therefore, important to reconstruct 3D objects that satisfy this property.

Previous works on generating physically sound 3D objects~\cite{mezghanni2022physical, ni2024phyrecon, yang2024physcene} have focused on specific object categories, such as furniture.
More recent methods like Atlas3D~\cite{chenatlas3d} and PhysComp~\cite{guo2025physically} tackle a broader range of object categories.
Both methods optimize a 3D model, either from scratch~\cite{chenatlas3d} or from the output of an off-the-shelf 3D generator~\cite{guo2025physically}, using \emph{differentiable} physics-based losses that reward stability.
To compute these losses, they require differentiable simulators such as~\cite{hu2019difftaichi, warp2022}, which, despite continuous improvements, remain slower and numerically less stable than non-differentiable simulators like~\cite{todorov2012mujoco, makoviychuk2021isaac}.
As a result, Atlas3D and PhysComp are slow and susceptible to local optima and numerical instability.

In this paper, we aim to improve a feed-forward 3D generator so that it \emph{directly} outputs physically stable objects without requiring test-time corrections.
A na{\"\i}ve approach would be to use losses similar to those proposed by Atlas3D and PhysComp for feed-forward training instead of test-time optimization, but this would still require a differentiable simulator.
Instead, inspired by works on aligning generative models with human preferences~\cite{rafailov2023direct, wallace2024diffusion}, we introduce \textbf{D}irect \textbf{S}imulation \textbf{O}ptimization (\emph{\method}).
This simple and effective approach fine-tunes a 3D generator by aligning it with the ``preference'' provided automatically by an off-the-shelf physics simulator.
With this, we explore three research questions:
(1) How to use this simulation preference dataset to fine-tune a 3D generator efficiently;
(2) How to construct such a dataset \emph{without} requiring ground-truth 3D data; and
(3) Whether the fine-tuned generator generalizes well, outputting physically sound 3D objects from image prompts unseen during training.

Our motivation for using reward optimization is that stability, like many other physical attributes of an object, is \emph{discrete}: either an object is stable, or it collapses under gravity.
Stability does not distinguish between unstable states regardless of how close they are to becoming stable, making it difficult to optimize using techniques like gradient descent.
In contrast, it is easy to determine whether an object is stable or not using a physics simulator.
Hence, we reformulate the problem as a \emph{reward-based learning task}, where we reward stable outputs and penalize unstable ones.
Inspired by direct preference optimization (DPO)~\cite{rafailov2023direct}, we propose an alternative objective, direct \emph{reward} optimization (DRO), for aligning diffusion models with external preferences.
Notably, DRO does not require \emph{pairwise} preference data for training.

Our second contribution is to show that we can derive reward signals solely from generated data, eliminating the need to collect large datasets of stable 3D objects for training at scale.
We achieve this by generating new 3D assets using the 3D generator itself.
These generated 3D assets are then evaluated within a physics simulator, classifying them as stable or unstable.
This process allows us to construct a fully automated self-improving pipeline, where the model is trained on its own output, assessed by a physics simulator rather than relying on a large dataset of 3D objects.

We show that, when integrated with either DPO or DRO as the objective function, our Direct Simulation Optimization framework can steer the output of the 3D generator to align with physical soundness.
The final model surpasses previous approaches for physically stable 3D generation on existing evaluation benchmarks~\cite{guo2025physically}.
It operates in a \emph{feed-forward} manner at test time, outperforming heavily engineered solutions like~\cite{chenatlas3d,guo2025physically} that perform test-time optimization, both in terms of speed and probability of generating a stable object as output.
The model also generalizes well to images collected in the wild (\cref{fig:teaser}).

Our experiments show that, in our setting, the proposed DRO objective achieves faster convergence and superior alignment compared to DPO, suggesting that it may be a better candidate for diffusion alignment in general.
While our study focuses on stability under gravity, the reward-based approach and the self-improving optimization strategy can, in principle, be applied to any physical attributes that can be assessed via a simulator.

\section{Related Work}%
\label{sec:rel}

\paragraph{3D generation and reconstruction.}

Early 3D generators used generative adversarial networks (GANs)~\cite{goodfellow2014generative} and various 3D representations such as point clouds~\cite{lin2018learning,huang2020pf}, voxel grids~\cite{wu2016learning,yang20173d,zhu2018visual}, view sets~\cite{park2017transformation,nguyen2019hologan}, NeRF~\cite{schwarz2020graf,chan2021pi,niemeyer2021giraffe,deng2022gram,chan2022efficient}, SDF~\cite{gao2022get3d}, and 3D Gaussian mixtures~\cite{wewer2024latentsplat}.
However, GANs are challenging to train on a large scale in an `open world' setting.
This explains why recent methods have shifted to diffusion models~\cite{sohl2015deep,ho2020denoising}, which use the same 3D representations~\cite{luo2021diffusion,nichol2022point,muller2023diffrf,shue20233d,Szymanowicz_2023_ICCV,chen2024mvsplat360} while improving training stability and scalability.
Other approaches train neural networks~\cite{kanazawa2018learning,ye2021shelf,wu2020unsupervised,yu2021pixelnerf,wu2023magicpony,jakab2024farm3d,li2024learning,huang2023shapeclipper,szymanowicz2024splatter,charatan2024pixelsplat,chen2024mvsplat,szymanowicz2024flash3d} to directly regress 3D models from 2D images.
Researchers have also explored scaling 3D reconstruction models~\cite{hong2023lrm,tang2025lgm,wang2024crm} on Objaverse~\cite{deitke22objaverse,deitke23objaverse-xl}, improving generalization.
DreamFusion~\cite{poole2022dreamfusion} and SJC~\cite{wang2023score} leverage large-scale image/video generators for 3D generation using score distillation~\cite{poole2022dreamfusion,wang2023score,lin2023magic3d,wang2023prolificdreamer,jakab2024farm3d,melas2023realfusion,zhu2024dreamhoi}.
The works of~\cite{liu2023zero,shi23mvdream,li2024instant3d,liu2023syncdreamer,melas-kyriazi2024IM3d,han2024vfusion3d,zheng2024free3d,weng2023consistent123,long2024wonder3d,gao2024cat3d,yang2024hunyuan3d} fine-tune these models for generalizable 3D generation.
More recently, researchers have introduced latent 3D representations~\cite{zhang20233dshape2vecset, xiang2024trellis, chen20243dtopia} whose distributions can be effectively modeled by denoising diffusion or rectified flow~\cite{albergo2023building, liu2023flow, lipman2023flow}.
CLAY~\cite{zhang2024clay} and TRELLIS~\cite{xiang2024trellis} are among the 3D generators trained in this manner, producing superior results compared to methods that rely on 2D generation.

These advances have significantly improved the quality of the geometry and appearance of generated 3D assets, but not necessarily their physical soundness.
This limitation reduces their utility in downstream applications like fabrication and simulation.
In contrast, we propose a 3D generation approach that explicitly optimizes physical soundness, specifically stability under gravity.

\paragraph{Physically-sound 3D generation.}

Early studies explored methods to predict physical properties from images and videos, such as mass~\cite{standley2017image2mass}, shadows~\cite{wang2021single}, materials~\cite{zhai2024physical}, occlusions~\cite{zhan2024does}, and support~\cite{silberman2012indoor}.
While effective in predicting specific physical parameters, these methods do not generalize directly to 3D reconstruction.
Recent works like Physdiff~\cite{yuan2023physdiff}, PhysGaussian~\cite{xie2024physgaussian}, and PIE-NeRF~\cite{feng2024pie} extend physics-based rendering~\cite{hu2019difftaichi} to NeRF~\cite{mildenhall2020nerf} and 3D Gaussian Splatting~\cite{kerbl20233d}.
These methods focus on modeling the motion of objects rather than their stability under gravity.
Similar to our work, Phys-DeepSDF~\cite{mezghanni2022physical}, PhyScene~\cite{yang2024physcene}, and PhyRecon~\cite{ni2024phyrecon} incorporate explicit physical constraints in 3D reconstruction.
However, these methods are limited to specific object categories, such as furniture.
More related to our work, Atlas3D~\cite{chenatlas3d} and PhysComp~\cite{guo2025physically} are not restricted to specific categories; instead, they rely on test-time optimization using carefully designed differentiable, physics-based losses.
We address a similar problem but in a \emph{feed-forward} manner using \emph{reward-based optimization}, avoiding the need for fragile and slow physics-based losses at test time.

\paragraph{Preference alignment in generative models.}

The Direct Simulation Optimization (DSO) framework we propose can be trained using Direct Preference Optimization (DPO)~\cite{rafailov2023direct}, a technique initially developed for fine-tuning large language models.
Diffusion-DPO~\cite{wallace2024diffusion} first extended DPO to vision diffusion models, enabling direct optimization of human preferences, and was further extended by~\cite{furuta2024improving,liu2025improving}.
While various preference alignment approaches exist~\cite{black2023training, fan2023dpok, prabhudesai2024video, ouyang2022training, lee2023aligning}, DPO has the distinct advantage of not requiring an oracle to compute the reward signal during training and avoids the need for reward modeling.
Inspired by DPO, we also propose an alternative objective named direct reward optimization (DRO), which does not require \emph{pairwise} preference data to align the generator.
\section{Method}%
\label{sec:method}

Given a pre-trained diffusion-based 3D generator $p_\text{ref}$ that takes a single image $I$ as input and generates 3D assets $\bm{x}_0 \sim p_\text{ref}(\bm{x}_0|I)$, our goal is to learn a new model $p_\theta$ that produces more physically sound generations than $p_\text{ref}$.
We assume access to an oracle $o$ that, given a sample $\bm{x}_0$, outputs $o(\bm{x}_0) \in \{0, 1\}$, indicating whether $\bm{x}_0$ is physically sound. 
In this paper, we focus on stability under gravity, where $o$ is computed by a physics simulator to determine whether a 3D model $\bm{x}_0$ is self-supporting.

\begin{figure*}[t!]
\centering
\includegraphics[width=\linewidth, clip, trim=0 5 0 0]{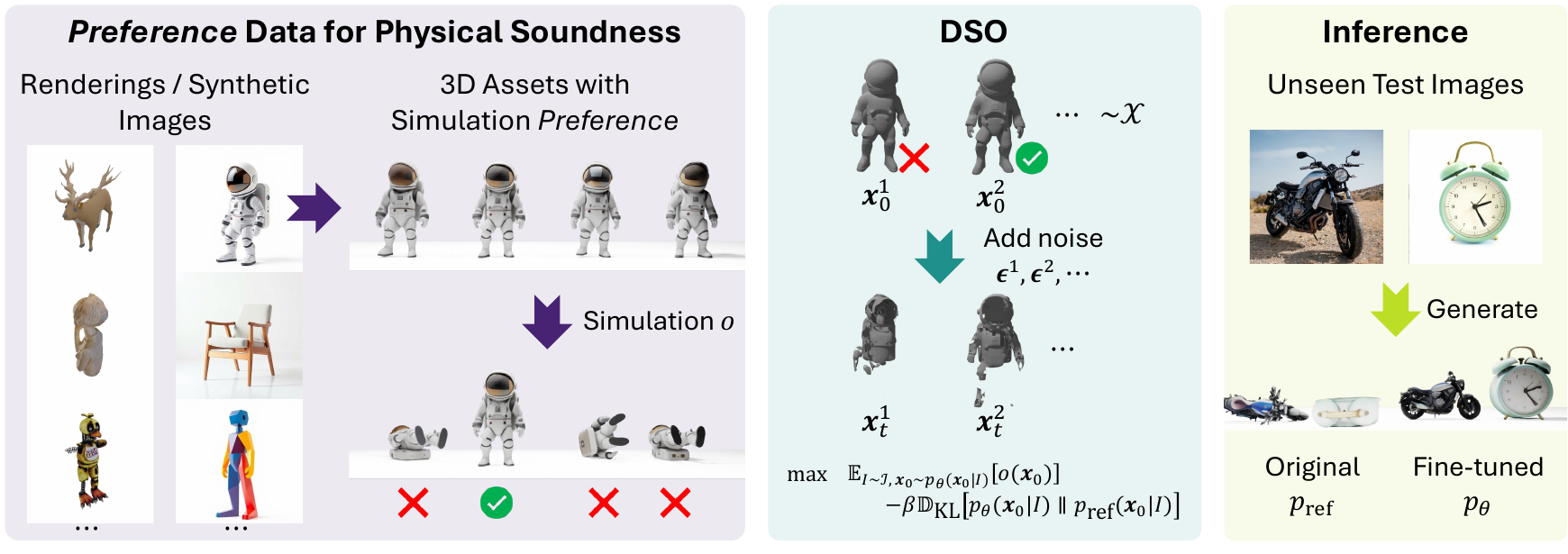}
\caption{\textbf{Overview} of \textbf{D}irect \textbf{S}imulation \textbf{O}ptimization (\textbf{DSO}).
\emph{Left}: Starting from a set of (potentially synthetic) image prompts, we task the base model $p_\text{ref}$ to generate 3D models. Each model is augmented with a binary stability label through physics-based simulation (\cref{sec:method_data}).
\emph{Middle}: Using this dataset, we fine-tune the base model by reinforcing stable samples and discouraging unstable ones. Our objective formulation enables efficient training via gradient descent without \emph{pairwise} preferences (\cref{sec:method_formulation}).
\emph{Right}: At test time, the fine-tuned model can generate self-supporting objects when conditioned on (out-of-distribution) images of stable objects captured \emph{in the wild}.
}%
\label{fig:method}
\end{figure*}

\subsection{Challenges of Optimizing Physical Soundness}%
\label{sec:method_objectives}

To improve the physical soundness of the generated samples, one approach is to fine-tune the model with the following objective:
\begin{align}
\label{eq:naive_objective}
    \max_{\theta} \mathbb{E}&_{I \sim \mathcal{I}, \bm{x}_0 \sim p_\theta(\bm{x}_0|I)}
    \left[o(\bm{x}_0)\right]  
    \notag \\ 
    & - \beta \mathbb{D}_\text{KL}\left[ p_\theta(\bm{x}_0|I) \Vert p_\text{ref}(\bm{x}_0|I) \right],
\end{align}
where $\mathcal{I}$ is the empirical distribution of a dataset of image prompts, and $\beta$ is a hyperparameter trading off the two terms.
The first term encourages the generated object $\bm{x}_0$ from $p_\theta(\bm{x}_0|I)$ to be physically sound, while the second term constrains the distribution to remain close to the base model to ensure that the generated geometry remains faithful to the input image $I$.

A key challenge in optimizing~\cref{eq:naive_objective} is that the oracle $o$ is \emph{non-differentiable}.
One approach to address this issue is to reframe the denoising process as a multi-step Markov decision process (MDP)~\cite{black2023training} and optimize~\cref{eq:naive_objective} using reinforcement learning (RL)~\cite{schulman2015trust, schulman2017proximal}.
However, in our setting, evaluating $o$ is computationally expensive due to the need to run a physical simulation and the overhead introduced by decoding latent 3D representations $\bm{x}_0$ into simulation-ready assets.
The decoding process of state-of-the-art 3D generators involves querying dense 3D grid points and extracting a 3D mesh with marching cubes~\cite{zhang2024clay, hunyuan3d22025tencent}, and may even require inference of another geometry generator~\cite{xiang2024trellis}.
These factors make optimization of~\cref{eq:naive_objective} via RL computationally prohibitive.

\subsection{Formulation as Reward Optimization}%
\label{sec:method_formulation}

We aim to reformulate the objective function to be easier to optimize, specifically eliminating the need to evaluate $o$ during training, while still preserving the intended goals of~\cref{eq:naive_objective}.
This is analogous to the goal of text-to-image diffusion model alignment in Diffusion-DPO~\cite{wallace2024diffusion}.
In both cases, the reward signal (\ie, evaluation of $o$ by simulation or by collecting human preferences for~\cite{wallace2024diffusion}) is hard to obtain in a scalable way during training.

Following Diffusion-DPO~\cite{wallace2024diffusion}, we can re-parameterize $o(\bm{x}_0)$ using the optimal reverse diffusion process, modeled by $p^\star_\theta(\bm{x}_{0:T})$, that maximizes (a lower bound of)~\cref{eq:naive_objective}:
\begin{equation}
\label{eq:reward}
    o(\bm{x}_{0}) = \beta \mathbb{E}_{p_\theta(\bm{x}_{1:T}|\bm{x}_0, I)}\left[ \log \frac{p^\star_\theta(\bm{x}_{0:T} | I)}{p_\text{ref}(\bm{x}_{0:T}|I)}\right] + \beta \log Z(I),
\end{equation}
for any $I\in \operatorname{supp}(\mathcal{I})$, where $Z(I)$ is a normalizing term independent of $p_\theta$.
The derivation follows~\cite{wallace2024diffusion} and is detailed in~\cref{sec:supp_derivation}.

\paragraph{Direct Reward Optimization (DRO).}

Given an image dataset $\mathcal{I}$ and 3D models $\mathcal{X}_I$ corresponding to each image $I\in \mathcal{I}$, we can pre-compute $o(\bm{x}_0)$ for each 3D model $\bm{x}_0 \in \mathcal{X}_I$ to supervise $p_\theta$ using~\cref{eq:reward}, via an $L1$ loss:
\begin{equation}
\label{eq:naive}
    \begin{aligned}
        \mathcal{L} \coloneqq &\mathbb{E}_{I\sim \mathcal{I}, \bm{x}_0\sim \mathcal{X}_I} 
        \bigg[ \bigg\vert o(\bm{x}_0) - \beta \bigg( \\
        & \mathbb{E}_{p_\theta(\bm{x}_{1:T}|\bm{x}_0, I)}\left[ \log \frac{p_\theta(\bm{x}_{0:T} | I)}{p_\text{ref}(\bm{x}_{0:T}|I)}\right]
        + \log Z(I)
        \bigg)\bigg\vert \bigg].
    \end{aligned}
\end{equation}
However, despite $\mathcal{L}$ being a function of the trainable parameters $\theta$, it is intractable because neither $Z(I)$ nor the expectation over $\bm{x}_{1:T}$ can be computed efficiently.

To address this issue, we notice that the absolute value of $o(\bm{x}_0)$ is arbitrary, \ie, we could use another oracle $o^\prime(\bm{x}_0) \in \{l, u\}$ which evaluates to $l$ for unstable $\bm{x}_0$ and $u$ for stable $\bm{x}_0$, as long as $l < u$ in~\cref{eq:naive_objective}.
In this setting, there exists a choice of $\beta$ that leads to the same optimum $p_\theta^\star$ as with the original oracle $o$ in~\cref{eq:naive_objective}.

Since we aim to use stochastic gradient descent, which is \emph{local} and \emph{continuous}, to optimize $\mathcal{L}$, we may as well choose $l$ and $u$ such that, within the training compute budget, the sum of $\log Z(I)$ and the expectation over $\bm{x}_{1:T}$ is bounded within $(\frac{l}{\beta}, \frac{u}{\beta})$.
By doing so, we can remove the absolute value in~\cref{eq:naive} and get rid of the terms independent of $p_\theta$:
\begin{equation}
\label{eq:simplify}
    \begin{aligned}
        \arg\min \mathcal{L} =
        \arg\min &\mathbb{E}_{I\sim \mathcal{I}, \bm{x}_0\sim \mathcal{X}_I, \bm{x}_{1:T}\sim p_\theta(\bm{x}_{1:T}|\bm{x}_0, I)}\bigg[ \\
        (1 - 2o(\bm{x}_0))&\log \frac{p_\theta(\bm{x}_{0:T} | I)}{p_\text{ref}(\bm{x}_{0:T}|I)}
        \bigg].
    \end{aligned}
\end{equation}

To make sampling tractable, we approximate the reverse process $p_\theta(\bm{x}_{1:T}|\bm{x}_0, I)$ with the forward process $q(\bm{x}_{1:T} | \bm{x}_0)$, following~\cite{wallace2024diffusion}.
With some algebra, this yields:
\begin{equation}
    \label{eq:dro_objective}
    \begin{aligned}
        \mathcal{L}_\text{DRO} =& -T \mathbb{E}_{I\sim \mathcal{I}, \bm{x}_0\sim \mathcal{X}_I, t\sim\mathcal{U}(0, T), \bm{x}_{t}\sim q(\bm{x}_t|\bm{x}_0)} \bigg[ \\
        &\quad w(t)(1 - 2o(\bm{x}_0)) \Vert \bm{\epsilon} - \bm{\epsilon}_\theta(\bm{x}_t, t) \Vert ^2_2\bigg],
    \end{aligned}
\end{equation}
where $\bm{\epsilon}\sim \mathcal{N}(0, \mathbf{I})$ is a draw from $q(\bm{x}_t|\bm{x}_0)$ and $w(t)$ is a weighting function.
$\mathcal{L}_\text{DRO}$ directly encourages the model to improve at denoising samples $\bm{x}_0$ with high reward (\ie, $o(\bm{x}_0)=1$) and to denoise less well samples $\bm{x}_0$ with low reward (\ie, $o(\bm{x}_0)=0$).
We hence dub it direct reward optimization (\emph{DRO}).
Different from the DPO formulation~\cite{wallace2024diffusion} (which we briefly review next),
fine-tuning with $\mathcal{L}_\text{DRO}$ does not require \emph{pairwise} preference data and does not query the base model $\bm{\epsilon}_\text{ref}$ during training, potentially applicable to more alignment settings than DPO\@.

\paragraph{Direct Preference Optimization (DPO).}
Alternatively, assuming $\mathcal{X}_I$ contains both stable and unstable models, we can use the objective introduced in~\cite{wallace2024diffusion}, which relies on \emph{pairwise} preference data and minimizes a \emph{contrastive} loss:
{
\thinmuskip=1mu
\medmuskip=2mu
\thickmuskip=3mu
\begin{equation}
\label{eq:dpo_objective_pre}
    \begin{aligned}
        \mathcal{L}_\text{DPO} \coloneqq &-\mathbb{E}_{I\sim \mathcal{I}, (\bm{x}_0^w, \bm{x}_0^l)\sim \mathcal{X}_I^2} 
        \bigg[\log \operatorname{sigmoid}(r(\bm{x}_0^w) - r(\bm{x}_0^l)) \bigg],
    \end{aligned}
\end{equation}
}
where $(\bm{x}_0^w, \bm{x}_0^l)$ is a pair of physically sound and unsound 3D models corresponding to the same image $I$ (\ie, $o(\bm{x}_0^w)=1-o(\bm{x}_0^l) = 1$), and $r$ is a reward model introduced to derive the loss from the Bradley-Terry model~\cite{bradley1952rank}.
Following the derivation in~\cite{wallace2024diffusion}, this simplifies to:

\begin{equation}
\label{eq:dpo_objective}
    \begin{aligned}
        \mathcal{L}_\text{DPO} ={}& -\mathbb{E}_{\substack{I\sim \mathcal{I}, (\bm{x}_0^w, \bm{x}_0^l)\sim \mathcal{X}_I^2, t\sim\mathcal{U}(0, T), \bm{x}_{t}^w\sim q(\bm{x}_t^w|\bm{x}_0^w), \bm{x}_{t}^l\sim q(\bm{x}_t^l|\bm{x}_0^l)}} \\
        &\log \operatorname{sigmoid} \bigg( -\beta T w(t)\Big(\\
        &\quad \Vert \bm{\epsilon}^w - \bm{\epsilon}_\theta(\bm{x}_t^w, t)\Vert ^2 _2 - \Vert \bm{\epsilon}^w - \bm{\epsilon}_\text{ref}(\bm{x}_t^w, t) \Vert^2_2 \\
        &\quad -\left( \Vert \bm{\epsilon}^l - \bm{\epsilon}_\theta(\bm{x}_t^l, t) \Vert ^2_2 - \Vert \bm{\epsilon}^l - \bm{\epsilon}_\text{ref}(\bm{x}_t^l, t) \Vert ^2_2 \right) \Big) \bigg) ,
    \end{aligned}
\end{equation}

where $\bm{\epsilon}^w, \bm{\epsilon}^l\sim\mathcal{N}(0, \mathbf{I})$ are two independent random draws.
Please refer to~\cite{wallace2024diffusion} for details.

\subsection{DSO with Generated Data}%
\label{sec:method_data}

We can now fine-tune the generator $p_\theta$\footnote{While our presentation in \cref{sec:method_formulation} focuses on a DDPM-formulated diffusion model with discrete timesteps~\cite{ho2020denoising}, the same approach can be readily adapted to rectified flow models~\cite{albergo2023building, liu2023flow, lipman2023flow} and other diffusion formulations~\cite{song2021scorebased, karras2022elucidating}, as their differences primarily lie in the noise schedule and loss weighting~\cite{gao2025diffusionmeetsflow}.}
with \cref{eq:dro_objective} or \cref{eq:dpo_objective} as the objective using stochastic gradient descent.
The final cornerstone of our framework, \textbf{D}irect \textbf{S}imulation \textbf{O}ptimization (\emph{\method}), is to obtain a set of images $\mathcal{I}$ and their corresponding 3D models $\mathcal{X}_{I\in \mathcal{I}}$.
Procuring a large number of stable 3D objects for training at scale is challenging, especially if we want multiple different objects for a single image prompt as in~\cref{eq:dpo_objective}.
Instead, we propose a scheme that leverages the 3D models generated by the generator $p_\text{ref}$ itself.
As illustrated in~\cref{fig:method}, we first curate a large, diverse image dataset $\mathcal{I}$.
These images can be either renderings of existing 3D datasets such as~\cite{deitke22objaverse, deitke23objaverse-xl}, or synthetic images generated by a 2D generator such as~\cite{rombach2022stablediffusion, flux2024}.
We then task the base model $p_\theta$ to create 3D models $\mathcal{X}_I$, taking individual images $I\in \mathcal{I}$ as input.
These 3D models, subsequently augmented with physical soundness scores via physics-based simulation, are used to fine-tune the model for enhanced physical soundness, achieving self-improvement \emph{without} relying on 3D ground truths.

\section{Experiments}%
\label{sec:exp}

We evaluate \method on the task of generating physically stable 3D models under gravity and compare it to prior works that use test-time optimization of physically-based losses (\cref{sec:exp_comp}).
We assess the ability to generate stable 3D objects while retaining the fidelity of the 3D reconstruction (as it would be trivial to make all objects stable by making them, \eg, cubes).
We discuss the effect of \method on the generated geometry in~\cref{sec:exp_geometry} and \method's scaling behavior in~\cref{sec:exp_scaling}.
In~\cref{sec:exp_no3d}, we demonstrate how \method can be adapted to leverage exclusively synthetic 2D images instead of renderings of ground-truth 3D models.

\subsection{Experiment Details}%
\label{sec:exp_details}

\paragraph{Model and data.}

We apply \method to fine-tune TRELLIS~\cite{xiang2024trellis}, a state-of-the-art image-to-3D generator, and measure its ability to consistently generate self-supporting 3D models before and after optimization.
TRELLIS contains \emph{two} rectified flow transformers: the first generates the coarse geometry of the 3D object, and the second refines its fine-grained details.
In our experiments, we fine-tune only the linear layers of the first transformer, as stability is primarily controlled by the coarse geometry.
We use LoRA~\cite{hu2022lora} to reduce the number of parameters to optimize.
We select TRELLIS because it is a state-of-the-art 3D generator and is available as open source, but our method is not specific to this model.

For the training data, we first generate a large number of 3D models with TRELLIS, conditioned on Objaverse~\cite{deitke22objaverse} renderings.
We exclude objects from Objaverse with unstable ground-truth shapes and filter out low-quality ones following~\cite{xiang2024trellis}.
Additionally, we include only objects categorized by GObjaverse~\cite{qiu2024richdreamer} as ``\texttt{Human-Shape}'', ``\texttt{Animals}'', or ``\texttt{Daily-Used}'', as these categories often feature two-legged shapes and tall, slender structures, making them more challenging to stabilize under gravity.
We render $6$ images for each of the remaining $13$k objects and generate $4$ different models per image, yielding $312$k 3D models in total.
We then use the MuJoCo~\cite{todorov2012mujoco} simulator to conduct physical simulations for each model, starting from an upright pose on flat ground.
We use the tilting angle at the final equilibrium state to determine stability, based on a hard cut-off of $20^\circ$: a model $\bm{x}_0$ is considered stable (\ie, $o(\bm{x}_0) = 1$) if its tilting angle is below $20^\circ$ and unstable otherwise.
During training, we sample models for an image prompt uniformly at random.

\paragraph{Training.}

We use AdamW~\cite{loshchilov2017decoupled} to fine-tune the base model using LoRA~\cite{hu2022lora} (rank $64$) with a batch size of $48$ on $4$ NVIDIA A100 GPUs.
We train \emph{two} separate models, optimizing them using $\mathcal{L}_\text{DRO}$ (\cref{eq:dro_objective}) for $4,000$ steps and using $\mathcal{L}_\text{DPO}$ (\cref{eq:dpo_objective}) for $8,000$ steps, respectively.
The $\beta$ in~\cref{eq:dpo_objective} is set to $500$.
More details can be found in~\cref{sec:supp_training_details}.

\paragraph{Evaluation.}

We evaluate on the dataset from~\cite{guo2025physically}, which consists of $100$ Objaverse~\cite{deitke22objaverse} objects from plants, animals, and characters. 
We exclude the $35$ objects whose ground-truth shape is \emph{not} self-supporting and render $12$ images for each of the remaining objects, resulting in a final set of $65$ objects and $780$ images.
These objects are removed from our training set.

\paragraph{Metrics.}

For quantitative results, we report the following stability measures:
\emph{$\%$ Output} counts the frequency of successfully outputting a 3D object, regardless of its stability;
\emph{$\%$ Stable} counts the percentage of stable assets among those generated;
\emph{Rotation angle} (\emph{Rot.} in short) measures the average tilting angle of generated objects at their equilibrium state.
In addition, to evaluate the mesh geometry, we report \emph{Chamfer Distance} (\emph{CD}) and \emph{F-Score} (with threshold $0.05$~\cite{melas2023realfusion, wang2024crm, liu2023one}).
Following common practices~\cite{melas2023realfusion, wang2024crm, liu2023one}, we scale the meshes to fit within the unit cube and align the generated meshes optimally with the ground truths using Iterated Closest Point (ICP) before computing CD and F-Score.

\begin{table}[tb!]
    \tablestyle{3.0pt}{1.0}
    \centering
    \begin{tabular}{@{}lcc cc@{}}
        \toprule
        \multirow{2}{*}{\textbf{Method}} & \multicolumn{2}{c}{\textbf{Stability}} & \multicolumn{2}{c}{\textbf{Geometry}} \\
        \cmidrule(lr){2-3}\cmidrule(lr){4-5}
        & \makecell{$\%$ Stable$\uparrow$ \\ ($\%$ Output$\uparrow$)} & Rot.$\downarrow$ & CD$\downarrow$ & F-Score$\uparrow$ \\
        \midrule
        \multicolumn{5}{@{}l}{Full evaluation set ($65$ objects)}  \\
        \cmidrule(lr){1-2}
        TRELLIS~\cite{xiang2024trellis} & $85.1$ ($\mathbf{100}$) & $14.14^\circ$ & $0.0485$ & $73.12$ \\
        Atlas3D~\cite{chenatlas3d} & $69.4$ ($95.4$) & $32.86^\circ$ & --- & --- \\
        \textbf{TRELLIS + \method(w/ $\mathcal{L}_\text{DPO}$)} & $\underline{95.1}$ ($\mathbf{100}$) & $\underline{5.42}^\circ$ & $\underline{0.0480}$ & $\underline{73.62}$\\
        \textbf{TRELLIS + \method(w/ $\mathcal{L}_\text{DRO}$)} & $\mathbf{99.0}$ ($\mathbf{100}$) & $\mathbf{1.88}^\circ$ & $\mathbf{0.0440}$ & $\mathbf{76.17}$ \\
        \midrule
        \multicolumn{5}{@{}l}{Partial evaluation set ($11$ unstable objects)} \\
        \cmidrule(lr){1-2}
        TRELLIS~\cite{xiang2024trellis} & $54.5$ ($\mathbf{100}$) & $39.18^\circ$ & $0.0529$ & $72.48$ \\
        TRELLIS + PhysComp~\cite{guo2025physically} & $80.3$ ($46.2$) & $18.14^\circ$ & $0.0698$ & $53.73$ \\
        \textbf{TRELLIS + \method(w/ $\mathcal{L}_\text{DPO}$)} & $\underline{82.6}$ ($\mathbf{100}$) & $\underline{16.83}^\circ$ & $\mathbf{0.0509}$ & $\underline{73.07}$ \\
        \textbf{TRELLIS + \method(w/ $\mathcal{L}_\text{DRO}$)} & $\mathbf{95.5}$ ($\mathbf{100}$) & $\mathbf{5.58}^\circ$ & $\underline{0.0520}$ & $\mathbf{73.61}$ \\
        \bottomrule
    \end{tabular}
    \caption{\textbf{Quantitative Results.} \method fine-tuned models (using either $\mathcal{L}_\text{DRO}$ or $\mathcal{L}_\text{DPO}$) significantly outperform baseline methods Atlas3D~\cite{chenatlas3d} and PhysComp~\cite{guo2025physically} in both physical stability and geometric quality.
    Beyond improving the physical soundness of the base model TRELLIS~\cite{xiang2024trellis}, \method also slightly improves its geometric fidelity \emph{without} requiring ground-truth 3D supervision.
    }
    \label{tab:comparison}
\end{table}
\begin{figure*}[tb!]
\centering
\includegraphics[width=\textwidth, clip, trim=0 10 0 0]{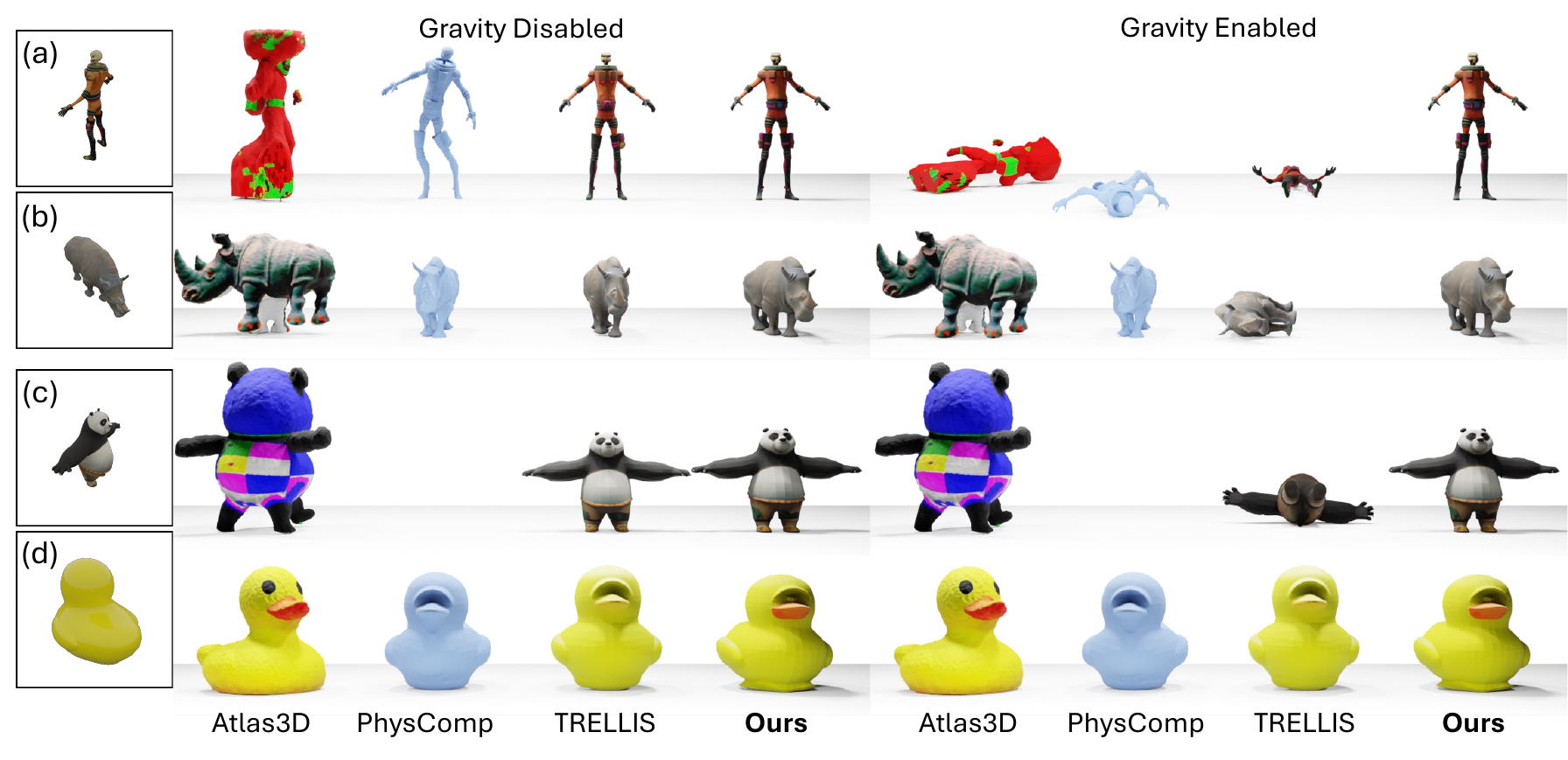}
\caption{\textbf{Qualitative Comparison} with baseline methods.
Our model can more reliably generate 3D assets that are stable under gravity and faithful to the conditioning images.
}%
\label{fig:qualitative-comparison}
\end{figure*}

\paragraph{Baselines.}

In addition to our base model TRELLIS~\cite{xiang2024trellis}, we consider two baseline methods designed to generate self-supporting 3D objects: \emph{Atlas3D}~\cite{chenatlas3d} and \emph{PhysComp}~\cite{guo2025physically}.
Atlas3D is a text-to-3D framework that combines score distillation sampling~\cite{poole2022dreamfusion} with physically-based loss terms, primarily the magnitude of the object orientation change at equilibrium, computed via differentiable simulation.
PhysComp takes a (volumetric) tetrahedral mesh as input and applies test-time optimization to improve its physical soundness, including its stability under gravity.
This is achieved by encouraging the projection of the center of mass to be within the convex hull of the contact points.
For~\cite{chenatlas3d} and~\cite{guo2025physically}, we use their official implementations.
For the text-conditioned Atlas3D, we prompt it using captions of our multi-view renderings, obtained with GPT-4V~\cite{openai23gpt4}.
We generate \emph{one} asset per object in the evaluation set.
For PhysComp, we task it to optimize the 3D models generated by TRELLIS\@.
Since the optimization on our hardware ($24$-core CPU with $668$ GiB RAM in total) takes significantly longer (on average $15$ minutes) than the $80$ seconds reported by the authors, we only run it on an $11$-object subset whose renderings lead TRELLIS to generate unstable 3D models, amounting to $11\times12=132$ runs.
As the optimization time varies dramatically with mesh complexity, we set a strict time budget of $30$ minutes per run.

\subsection{Results and Analysis}%
\label{sec:exp_comp}

\paragraph{Quantitative results.}

\Cref{tab:comparison} reports the quantitative results evaluated for both baselines and our method.
Notably, our \method fine-tuned TRELLIS (using either $\mathcal{L}_\text{DRO}$ or $\mathcal{L}_\text{DPO}$) outperforms all baselines on both physical stability and geometry fidelity \emph{without} any test-time optimization.

\paragraph{Qualitative results.}

\Cref{fig:qualitative-comparison} presents qualitative comparisons with baselines, highlighting cases where our base model TRELLIS~\cite{xiang2024trellis} fails to generate self-supporting assets.
Atlas3D~\cite{chenatlas3d}, inheriting the limitations of SDS-based approaches~\cite{poole2022dreamfusion}, often suffers from over-saturation and over-smoothness (a, b, c).
While incorporating physics-based stability loss, its optimization remains unreliable (a) and can introduce structural artifacts such as extraneous limbs (b, c).
PhysComp~\cite{guo2025physically}, which refines TRELLIS outputs, does not preserve texture and can distort the original shape (a), compromising faithfulness to the input image.
The method struggles to stabilize meshes in challenging scenarios (a) and frequently suffers from numerical instabilities, sometimes failing to generate outputs entirely (c).
In contrast, our final model leverages the strong geometric prior of TRELLIS while significantly enhancing physical stability without introducing additional computational overhead at test time.

\paragraph{Analysis.}

We note that:
(1) Differentiable simulation often suffers from numerical issues, as reflected by the lower $\%$ Output of~\cite{chenatlas3d} and~\cite{guo2025physically}, due to the need for differentiable ODE solving.
\method circumvents this requirement by framing physical soundness optimization as a reward learning task (\cref{sec:method_formulation}) and augmenting 3D models with simulation feedback before training.
(2) Unlike visual quality, physical stability demands high accuracy, especially in the contact region.
While existing efforts to align vision generators~\cite{wallace2024diffusion, furuta2024improving, liu2025improving, zhou2025dreamdpo, prabhudesai2024video} focus on enhancing visual quality, we show that alignment can also substantially improve accuracy-sensitive metrics.
(3) For our task, $\mathcal{L}_\text{DRO}$ proves to be a more effective objective than $\mathcal{L}_\text{DPO}$ (\cref{tab:comparison}) and could also be beneficial in other diffusion alignment settings, especially when access to \emph{pairwise} preference data is limited.

\begin{figure}[tb!]
\centering
\includegraphics[width=\columnwidth, clip, trim=0 20 0 0]{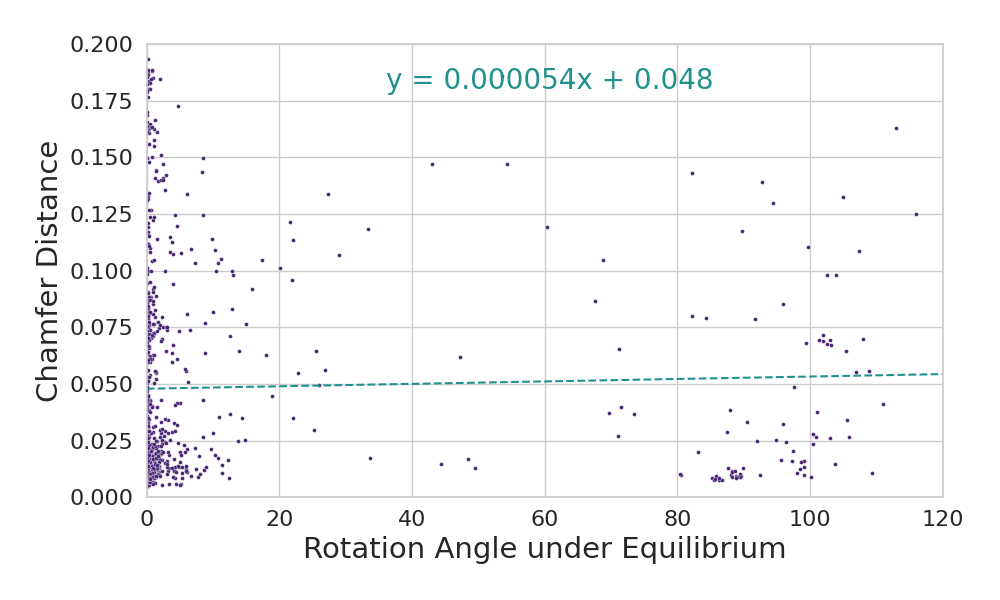}
\caption{\textbf{(Lack of) Correlation} between geometry reconstruction quality (Chamfer Distance) and stability (rotation angle).
}%
\label{fig:geometry-correlation}
\end{figure}

\subsection{Physical Soundness vs. Geometry Quality}%
\label{sec:exp_geometry}

\cref{eq:naive_objective} and the other losses in~\cref{sec:method} imply a potential trade-off between physical soundness and geometric quality, controlled by the parameter $\beta$.
However, in~\cref{tab:comparison}, \method fine-tuned on TRELLIS not only enhances physical stability but also improves \emph{geometric fidelity}.
This outcome is somewhat surprising, given that TRELLIS was explicitly \emph{trained} on (at least some) objects in the evaluation with geometric losses (\eg, occupancy), whereas our \method does \emph{not} directly supervise the base model with ground-truth geometry.

To investigate this, we generate $800$ distinct 3D assets using TRELLIS and analyze the relationship between their geometric quality (measured by CD) and physical stability (quantified by the tilting angle at equilibrium), as shown in \cref{fig:geometry-correlation}.
The correlation is not statistically significant, suggesting that improving physical soundness does not need to compromise geometric quality.
If anything, there is a very slight positive correlation between the two.

\begin{table}[tb!]
    \tablestyle{4.0pt}{1.0}
    \centering
    \begin{tabular}{@{}lcc cc@{}}
        \toprule
        \multirow{2}{*}{\textbf{Method}} & \multicolumn{2}{c}{\textbf{Stability}} & \multicolumn{2}{c}{\textbf{Geometry}} \\
        \cmidrule(lr){2-3}\cmidrule(lr){4-5}
        & $\%$ Stable$\uparrow$ & Rot.$\downarrow$ & CD$\downarrow$ & F-Score$\uparrow$ \\
        \midrule
        TRELLIS~\cite{xiang2024trellis} & $85.1$ & $14.14^\circ$ & $0.0485$ & $73.12$ \\
        TRELLIS + SFT & $89.5$ & $10.22^\circ$ & {$\mathbf{0.0440}$} & {$\mathbf{76.17}$} \\
        TRELLIS + \method w/ $\mathcal{L}_\text{DPO}$ & $\underline{95.1}$ & $\underline{5.42}^\circ$ & ${0.0480}$ & ${73.62}$ \\
        TRELLIS + \method w/ $\mathcal{L}_\text{DRO}$ & $\mathbf{99.0}$ & $\mathbf{1.88}^\circ$ & ${\mathbf{0.0440}}$ & ${\mathbf{76.17}}$ \\
        \bottomrule
    \end{tabular}
    \caption{\textbf{Comparison} with Supervised Fine-tuning (SFT).
    SFT yields faithful geometry, but its samples are less physically stable.
    }
    \label{tab:compare-sft}
\end{table}

\begin{figure*}[tb!]
    \centering
    \subfloat[
    \textbf{Scaling with training compute}.
    Longer training enhances the physical stability further, yet excessive training degrades geometric quality.
    \label{fig:scaling_compute}
    ]{
        \begin{minipage}{0.45\linewidth}
            \includegraphics[width=\linewidth, clip, trim=0 0 0 0]{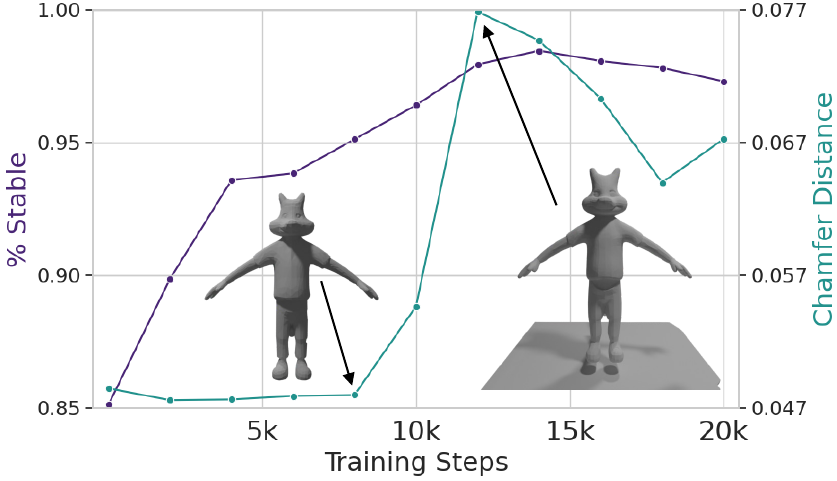}
        \end{minipage}
    }\hspace{2em}
    \subfloat[
        \textbf{Scaling with training data}.
        Smaller ($\frac{1}{16}$ of the full dataset) data achieve comparable results for physical alignment.
        \label{fig:scaling_data}
        ]{
            \begin{minipage}{0.45\linewidth}
                \includegraphics[width=\linewidth, clip, trim=0 0 0 0]{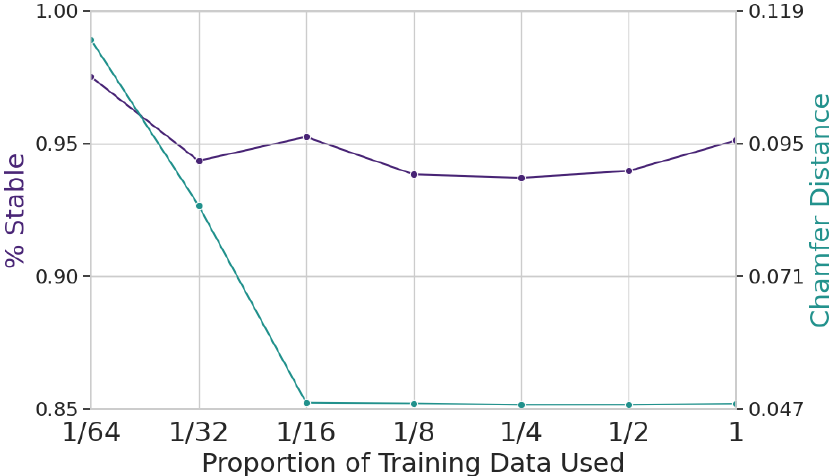}
            \end{minipage}
    }
    \vspace{-1em}
    \caption{\textbf{Scaling Behaviors} of \method with training compute (\emph{left}) and data (\emph{right}).
    }%
    \label{fig:scaling-compute}
\end{figure*}

\subsection{Comparison with Supervised Fine-tuning}%
\label{sec:exp_ablations}

To further assess the effectiveness of \method, we also compare it with supervised fine-tuning (SFT) in~\cref{tab:compare-sft}.
For SFT, we fine-tune TRELLIS on the stable subset of our constructed dataset (\ie, $\{\bm{x}_0\in\mathcal{X} | o(\bm{x}_0)=1\}$, consisting of $72$k objects out of the $312$k generated in total), using the rectified flow objective~\cite{albergo2023building, lipman2023flow, liu2023flow} with the same hyperparameter configuration as our main training runs for 8,000 steps.
While SFT yields better geometry, its samples are less physically sound.
This suggests that the model prioritizes geometry over physical plausibility, making fine-tuning 3D generators solely on physically stable objects less effective for aligning physical soundness.
In contrast, by exposing the model to both stable and unstable objects, \method encourages the model to better focus on physical properties.

\subsection{Scaling Behaviors}%
\label{sec:exp_scaling}

We study how \method scales when optimizing $\mathcal{L}_\text{DPO}$.

\paragraph{Scaling with training compute.}

\cref{fig:scaling_compute} illustrates the progression of evaluation metrics throughout training.
While longer training further enhances the physical stability measure, excessive training with \method significantly degrades geometric quality.
In particular, the model eventually ``cheats'' by generating a flat structure beneath the 3D asset as a base to prevent it from toppling over.

\paragraph{Scaling with training data.}

In~\cref{fig:scaling_data}, we analyze the impact of training data size on model performance.
We train $6$ models with identical hyperparameters as our main training run, progressively reducing the amount of data exposed to each model.
The smallest dataset used is only $\frac{1}{64}$ of the full dataset, constructed as described in~\cref{sec:exp_details}.
While training on extremely small datasets leads to model collapse, we find that using just $\frac{1}{16}$ of the full dataset (equivalent to $19.2$k synthetic 3D models with simulation feedback) already produces results comparable to our main training run.
This suggests that aligning state-of-the-art 3D generators with physical soundness requires only a modest amount of preference data.
This is promising for aligning other physical properties, such as 3D scene decomposition~\cite{yao2025cast, yang2024physcene, ni2024phyrecon} and part articulation~\cite{li2024dragapart, li2024puppetmaster, luo2024physpart}, for which obtaining positive samples may be more challenging due to their rarity.

\begin{table}[tb!]
    \tablestyle{2.0pt}{1.0}
    \centering
    \begin{tabular}{@{}lcccc cc@{}}
        \toprule
        \multirow{2}{*}{\textbf{Method}} & \multirow{2}{*}{\textbf{Synth.}} & \multirow{2}{*}{\textbf{Loss}} & \multicolumn{2}{c}{\textbf{Stability}} & \multicolumn{2}{c}{\textbf{Geometry}} \\
        \cmidrule(lr){4-5}\cmidrule(lr){6-7}
        & & & $\%$ Stable$\uparrow$ & Rot.$\downarrow$ &  CD$\downarrow$ & F-Score$\uparrow$ \\
        \midrule
        TRELLIS~\cite{xiang2024trellis} & --- & --- & $85.1$ & $14.14^\circ$ &  $0.0485$ & $73.12$ \\
        TRELLIS + \method & \checkmark & $\mathcal{L}_\text{DPO}$  & $93.5$ & $6.92^\circ$ & $0.0483$ & $73.40$ \\
        TRELLIS + \method & \ding{55} & $\mathcal{L}_\text{DPO}$ & $95.1$ & $5.42^\circ$ & $0.0480$ & $73.62$ \\
        TRELLIS + \method & \checkmark & $\mathcal{L}_\text{DRO}$ & \underline{$97.6$} & $\underline{3.17}^\circ$ & \underline{$0.0455$} & \underline{$76.05$} \\
        TRELLIS + \method & \ding{55} & $\mathcal{L}_\text{DRO}$ & $\mathbf{99.0}$ & $\mathbf{1.88}^\circ$ &  $\mathbf{0.0440}$ & $\mathbf{76.17}$ \\
        \bottomrule
    \end{tabular}
    \vspace{-0.5em}
    \caption{\method can be trained solely on \emph{synthetic} data. The resulting models achieve greater physical soundness than the base model.}
    \label{tab:synthetic}
\end{table}

\subsection{\method without Real Data}%
\label{sec:exp_no3d}

Our training objective does \emph{not} rely on ground-truth 3D data for supervision.
Nevertheless, in our main experiments presented in~\cref{sec:exp_comp}, we used Objaverse renderings as prompts to construct a preference dataset.
Here, we show that access to Objaverse models is \emph{not} necessary.
We substitute the renderings with object-centric synthetic images to condition the base model TRELLIS to generate 3D models.
We then evaluate the physical stability of these generated models using simulation feedback, assigning a binary preference label, which we use for \method fine-tuning.
In more detail, we task GPT-4~\cite{openai23gpt4} to generate 1,000 diverse prompts of detailed object descriptions and use them to prompt FLUX~\cite{flux2024}, an open-source text-to-image model, to generate synthetic images.
We then obtain a total of $64$k generated 3D assets, on which we conduct physical simulation as detailed in \cref{sec:exp_details}.
The performance of the model trained on this dataset is reported in~\cref{tab:synthetic}.
Despite the larger domain gap, the fine-tuned model generalizes well to the evaluation images and is more likely than the base model TRELLIS to generate stable assets under gravity.

\section{Conclusion}%
\label{sec:conclusion}

We presented \method, a novel framework for generating physically sound 3D objects by leveraging feedback from a physics simulator.
Our approach utilizes a dataset of 3D objects labeled with stability scores obtained from the simulator, potentially starting from entirely synthetic images.
We fine-tune the base generator using the DPO or DRO objectives, the latter of which we introduced.
The resulting \emph{feed-forward} generator is significantly faster and more reliable at producing stable objects compared to test-time optimization methods.

\paragraph{Acknowledgments.}

This work is supported by a Toshiba Research Studentship, EPSRC
SYN3D EP/Z001811/1, and ERC-CoG UNION 101001212.
We thank Minghao Guo and Bohan Wang for providing us with the evaluation set in their work~\cite{guo2025physically}, and Mariem Mezghanni for insightful discussions during the early stages of this project.
We also thank Zeren Jiang, Minghao Chen, Jinghao Zhou, and Gabrijel Boduljak for helpful suggestions.

{
\small
\bibliographystyle{ieee_fullname}
\bibliography{main}
}

\clearpage
\onecolumn

\appendix
\setcounter{section}{0} 
\renewcommand{\thesection}{\Alph{section}}

\section{Details of the Derivations}%
\label{sec:supp_derivation}

\paragraph{From \cref{eq:naive_objective} to \cref{eq:reward}.}
As in~\cite{wallace2024diffusion}, we introduce a latent oracle $O$ defined on the whole denoising chain $\bm{x}_{0:T}$, such that:
\begin{equation}
\label{eq:latent_oracle}
    o(\bm{x}_0) = \mathbb{E}_{p_\theta(\bm{x}_{1:T} | \bm{x}_0)}\left[O(\bm{x}_{0:T})\right].
\end{equation}
Then, starting from~\cref{eq:naive_objective}, we have:
\begin{equation}
\label{eq:detailed_derivation1}
    \begin{aligned}
            &\max_{\theta} \mathbb{E}_{I \sim \mathcal{I}, \bm{x}_0 \sim p_\theta(\bm{x}_0|I)}
            \left[o(\bm{x}_0)\right] - \beta \mathbb{D}_\text{KL}\left[ p_\theta(\bm{x}_0|I) \Vert p_\text{ref}(\bm{x}_0|I) \right] \\
            \geq &\max_{\theta} \mathbb{E}_{I \sim \mathcal{I}, \bm{x}_0 \sim p_\theta(\bm{x}_0|I)}
            \left[o(\bm{x}_0)\right] - \beta \mathbb{D}_\text{KL}\left[ p_\theta(\bm{x}_{0:T}|I) \Vert p_\text{ref}(\bm{x}_{0:T}|I) \right] \\
            = &\max_{\theta} \mathbb{E}_{I \sim \mathcal{I}, \bm{x}_{0:T} \sim p_\theta(\bm{x}_{0:T}|I)}
            \left[O(\bm{x}_{0:T})\right] - \beta \mathbb{D}_\text{KL}\left[ p_\theta(\bm{x}_{0:T}|I) \Vert p_\text{ref}(\bm{x}_{0:T}|I) \right] \\
            = &\beta \max_\theta \mathbb{E}_{I\sim \mathcal{I}, \bm{x}_{0:T}\sim p_\theta(\bm{x}_{0:T}|I)}\left[
            \log Z(I) - \log \frac{p_\theta(\bm{x}_{0:T}|I)}{p_\text{ref}(\bm{x}_{0:T}|I)\exp (O(\bm{x}_{0:T})/\beta)/Z(I)}
            \right],
    \end{aligned}
\end{equation}
where $Z(I) = \sum_{\bm{x}_{0:T}} p_\text{ref}(\bm{x}_{0:T}|I)\exp (O(\bm{x}_{0:T})/\beta)$ is a normalizing factor independent of $\theta$.
Since
\begin{equation}
    \mathbb{E}_{I\sim \mathcal{I}, \bm{x}_{0:T}\sim p_\theta(\bm{x}_{0:T}|I)}\left[\log \frac{p_\theta(\bm{x}_{0:T}|I)}{p_\text{ref}(\bm{x}_{0:T}|I)\exp (O(\bm{x}_{0:T})/\beta)/Z(I)}
            \right]
    = \mathbb{D}_\text{KL}\left[ p_\theta(\bm{x}_{0:T}|I) \Vert p_\text{ref}(\bm{x}_{0:T}|I)\exp(O(\bm{x}_{0:T})/\beta)/Z(I) \right] \geq 0
\end{equation}
with equality if and only if the two distributions are identical, the optimal $p^\star_\theta(\bm{x}_{0:T}|I)$ of the right-hand side of~\cref{eq:detailed_derivation1} has a unique closed-form solution:
\begin{equation}
    p^\star_\theta(\bm{x}_{0:T}|I) = p_\text{ref}(\bm{x}_{0:T}|I)\exp(O(\bm{x}_{0:T})/\beta)/Z(I).
\end{equation}
Therefore,
\begin{equation}
\label{eq:penultimate}
    O(\bm{x}_{0:T}) = \beta \log Z(I) + \beta \log \frac{p^\star_\theta(\bm{x}_{0:T}|I)}{p_\text{ref}(\bm{x}_{0:T}|I)}
\end{equation}
for any $I\in \operatorname{supp}(\mathcal{I})$.

We can then obtain~\cref{eq:reward} by plugging~\cref{eq:penultimate} into \cref{eq:latent_oracle}.

\paragraph{From \cref{eq:simplify} to \cref{eq:dro_objective}.}
Since sampling from $p_\theta(\bm{x}_{1:T}|\bm{x}_0, I)$ is intractable, we follow~\cite{wallace2024diffusion} and replace it with $q(\bm{x}_{1:T}|\bm{x}_0)$:
\begin{equation}
\label{eq:detailed_derivation2}
    \begin{aligned}
        \mathcal{L}_\text{DRO}\coloneqq&\min \mathbb{E}_{I\sim \mathcal{I}, \bm{x}_0\sim \mathcal{X}_I, \bm{x}_{1:T}\sim q(\bm{x}_{1:T}|\bm{x}_0)}\left[
        (1 - 2o(\bm{x}_0))\log \frac{p_\theta(\bm{x}_{0:T} | I)}{p_\text{ref}(\bm{x}_{0:T}|I)}
        \right] \\
        =& \min \mathbb{E}_{I\sim \mathcal{I}, \bm{x}_0\sim \mathcal{X}_I, \bm{x}_{1:T}\sim q(\bm{x}_{1:T}|\bm{x}_0)}\left[
        (1 - 2o(\bm{x}_0))\sum_{t=1}^{T}\log \frac{p_\theta(\bm{x}_{t-1} | \bm{x}_{t}, I)}{p_\text{ref}(\bm{x}_{t-1}| \bm{x}_{t}, I)}
        \right] \\
        =& \min T\mathbb{E}_{I\sim \mathcal{I}, \bm{x}_0\sim \mathcal{X}_I, t\sim\mathcal{U}(0, T), \bm{x}_{t}\sim q(\bm{x}_{t}|\bm{x}_0), \bm{x}_{t-1}\sim q(\bm{x}_{t-1}|\bm{x}_0, \bm{x}_t)}\left[
        (1 - 2o(\bm{x}_0))\log \frac{p_\theta(\bm{x}_{t-1} | \bm{x}_{t}, I)}{p_\text{ref}(\bm{x}_{t-1}| \bm{x}_{t}, I)}
        \right] \\
        =& \min T\mathbb{E}_{I\sim \mathcal{I}, \bm{x}_0\sim \mathcal{X}_I, t\sim\mathcal{U}(0, T), \bm{x}_{t}\sim q(\bm{x}_{t}|\bm{x}_0)}\bigg[
        (1 - 2o(\bm{x}_0)) \bigg(\\
        &\mathbb{D}_\text{KL}\left[ q(\bm{x}_{t-1}|\bm{x}_t, \bm{x}_0) \Vert p_\theta(\bm{x}_{t-1}|\bm{x}_t, I) \right] -
        \mathbb{D}_\text{KL}\left[ q(\bm{x}_{t-1}|\bm{x}_t, \bm{x}_0) \Vert p_\text{ref}(\bm{x}_{t-1}|\bm{x}_t, I) \right]
        \bigg)
        \bigg].
    \end{aligned}
\end{equation}
Recall that for diffusion models $p_\theta$ and $p_\text{ref}$, the distributions $q(\bm{x}_{t-1}|\bm{x}_t, \bm{x}_0)$, $p_\theta(\bm{x}_{t-1}|\bm{x}_t, I)$ and $p_\text{ref}(\bm{x}_{t-1}|\bm{x}_t, I)$ are all Gaussian.
Therefore, the KL divergence on the right-hand side of \cref{eq:detailed_derivation2} can be re-parameterized analytically using $\bm{\epsilon}_\theta$. 
After some algebra, and removing all terms independent of $\theta$, this yields \cref{eq:dro_objective}.

\section{Additional Training Details}%
\label{sec:supp_training_details}

\begin{table}[t] 
    \centering
    \noindent
    \begin{minipage}{0.49\textwidth}
        \tablestyle{3.0pt}{1.0}
        \centering
        \begin{tabular}{@{}lcc@{}}
            \toprule
            \textbf{Loss formulation}  & $\mathcal{L}_\text{DRO}$ & $\mathcal{L}_\text{DPO}$ \\
            \midrule
            \textbf{Optimization} & & \\
            Optimizer & AdamW & AdamW \\
            Learning rate & $5\times 10^{-6}$ & $5\times 10^{-6}$ \\
            Learning rate warmup & \makecell{ Linear \\ $2,000$ iterations } & \makecell{ Linear \\ $2,000$ iterations } \\
            Weight decay & $0.01$ & $0.01$ \\
            Effective batch size & $48$ & $48$ \\
            Training iterations & $4,000$ & $8,000$ \\
            Precision & \texttt{bf16} & \texttt{bf16} \\
            \midrule
            \textbf{LoRA} & & \\
            Rank & $64$ & $64$ \\
            $\alpha$ & $128$ & $128$ \\
            Dropout & $0$ & $0$ \\
            \midrule
            \textbf{Miscellaneous} & & \\
            Rectified flow $t$ sampling & $\operatorname{LogitNorm}(1, 1)$ & $\operatorname{LogitNorm}(1, 1)$ \\
            $\beta$ in $\mathcal{L}_\text{DPO}$ & --- & $500$ \\
            \bottomrule
        \end{tabular}
        \captionof{table}{\method training details and hyperparameter settings.}
        \label{tab:hyperparams}
    \end{minipage}
    \hfill
    \begin{minipage}{0.49\textwidth}
        \tablestyle{4.0pt}{1.0}
        \centering
        \begin{tabular}{@{}lcc cc@{}}
            \toprule
            \multirow{2}{*}{\textbf{Method}} & \multicolumn{2}{c}{\textbf{Alarm clock}} & \multicolumn{2}{c}{\textbf{Motorcycle}} \\
            \cmidrule(lr){2-3}\cmidrule(lr){4-5}
            & $\%$ Stable$\uparrow$ & Rot.$\downarrow$ & $\%$ Stable$\uparrow$ & Rot.$\downarrow$  \\
            \midrule
            TRELLIS~\cite{xiang2024trellis} & $67.5$ & $14.14^\circ$ & $44.4$ & $46.53^\circ$ \\
            TRELLIS + \method & $\mathbf{85.0}$ & $\mathbf{5.58}^\circ$ & ${\mathbf{58.1}}$ & ${\mathbf{36.75}^\circ}$ \\
            \bottomrule
        \end{tabular}
        \captionof{table}{\method enhances the model's ability to generate assets that remain stable under gravity from in-the-wild images of stable objects.}
        \label{tab:compare-real}
    \end{minipage}
\end{table}
All hyperparameters are listed in~\cref{tab:hyperparams}.
We did \emph{not} extensively tune these parameters: the LoRA parameters and the $\beta$ used in $\mathcal{L}_\text{DPO}$ follow~\cite{liu2025improving}, and the rectified flow noise level $t$ sampling uses the distribution from TRELLIS~\cite{xiang2024trellis}.

\section{Additional Evaluation Details}
\label{sec:supp_eval_details}

For evaluation, the 3D models are generated by TRELLIS~\cite{xiang2024trellis} and \method fine-tuned TRELLIS using the default setting:
$12$ sampling steps in stage 1 with classifier-free guidance $7.5$ and $12$ sampling steps in stage 2 with classifier-free guidance $3$.
Under this setting, generating \emph{one} model takes $10$ seconds on average on an NVIDIA A100 GPU.
By contrast, Atlas3D~\cite{chenatlas3d} takes $2$ hours to generate a model using SDS and PhysComp~\cite{guo2025physically} takes on average $15$ minutes to optimize \emph{one} model output by TRELLIS on our hardware.

We use MuJoCo~\cite{todorov2012mujoco} for rigid body simulation for evaluation.
The 3D models are assumed to be rigid and uniform in density.
We run the simulation for $10$ seconds, at which almost all objects have reached the steady state.

\section{Additional Results}

\subsection{Additional Evaluation Results}
To demonstrate that the enhanced physical soundness achieved through DSO is not limited to a specific simulation environment, we report the evaluation results in Isaac Gym~\cite{makoviychuk2021isaac} and under perturbations in~\cref{tab:additional-eval}.
For the evaluation under perturbations, we choose $4$ maximum perturbation angles $\theta_{\max}$ and perform $100$ simulation runs with each $\theta_{\max}$ where the generated 3D models are initially rotated by a random angle $\theta \in ( -\theta_{\max}, \theta_{\max} )$, following Atlas3D~\cite{chenatlas3d}.
We then report the average stability rate of the $100$ runs.
In~\cref{tab:additional-eval}, TRELLIS post-trained with only MuJoCo feedback via DSO outperforms all baselines under all simulation settings,
showing that the improved physical soundness generalizes well to different simulation environments.

\begin{table}[h!]
\small
\tablestyle{3.0pt}{1.0}
\setlength\tabcolsep{3pt}
\setlength{\abovecaptionskip}{2pt}
\renewcommand{\arraystretch}{1.0}
\centering
\begin{tabular}{@{}lcccccc@{}}
\toprule
\multirow{2}{*}{\textbf{Method}} &
\multicolumn{5}{c}{\textbf{MuJoCo}} &
\multicolumn{1}{c}{\textbf{Isaac Gym}} \\[-0.25ex]
\cmidrule(lr){2-6}\cmidrule(lr){7-7}
 & w/o perturbation & $\theta_{\max}=0.01$ & $\theta_{\max}=0.02$ & $\theta_{\max}=0.04$ & $\theta_{\max}=0.08$ & w/o perturbation \\
\midrule
\multicolumn{7}{@{}l}{\textit{Full evaluation set} ($65$ objects)}\\
\cmidrule(lr){1-1}
TRELLIS~\cite{xiang2024trellis}                         & $85.1$ & $84.8$ & $84.2$ & $82.5$ & $77.2$ & $97.3$ \\
Atlas3D~\cite{chenatlas3d}                          & $69.4$ & $70.3$ & $70.2$ & $66.3$ & $61.8$ & $88.7$ \\
\textbf{TRELLIS + \method(w/ $\mathcal{L}_\text{DPO}$)}
                                     & $\underline{95.1}$ & $\underline{94.8}$ & $\underline{94.1}$ & $\underline{92.6}$ & $\underline{88.0}$ & $\underline{99.3}$ \\
\textbf{TRELLIS + \method(w/ $\mathcal{L}_\text{DRO}$)}
                                     & $\mathbf{99.0}$ & $\mathbf{98.8}$ & $\mathbf{98.6}$ & $\mathbf{97.2}$ & $\mathbf{93.7}$ & $\mathbf{99.6}$ \\
\midrule
\multicolumn{7}{@{}l}{\textit{Partial evaluation set} ($11$ unstable objects)}\\
\cmidrule(lr){1-1}
TRELLIS~\cite{xiang2024trellis}                         & $54.5$ & $54.0$ & $53.8$ & $48.5$ & $41.5$ & $93.9$ \\
TRELLIS + PhysComp~\cite{guo2025physically}              & $80.3$ & $76.9$ & $76.1$ & $72.6$ & $\underline{67.7}$ & $83.9$ \\
\textbf{TRELLIS + \method(w/ $\mathcal{L}_\text{DPO}$)}
                                     & $\underline{82.6}$ & $\underline{82.0}$ & $\underline{80.7}$ & $\underline{77.5}$ & $67.5$ & $\underline{98.5}$ \\
\textbf{TRELLIS + \method(w/ $\mathcal{L}_\text{DRO}$)}
                                     & $\mathbf{95.5}$ & $\mathbf{95.4}$ & $\mathbf{95.0}$ & $\mathbf{93.9}$ & $\mathbf{85.4}$ & $\mathbf{100.0}$ \\
\bottomrule
\end{tabular}
\caption{\textbf{Results} evaluated under different simulation settings.}%
\label{tab:additional-eval}
\end{table}

\subsection{Additional Comparison with Post-Processing Baselines}
In~\cref{tab:additional-comp}, we compare DSO with a naive post-processing baseline that cuts the mesh flat just above the lowest vertex, following Atlas3D~\cite{chenatlas3d}.
This method is less effective at stabilizing meshes and significantly degrades geometric quality, as reflected in the higher Chamfer distance (\cref{tab:additional-comp}).

\begin{table}[ht!]
\small
\tablestyle{3.0pt}{1.0}
\setlength\tabcolsep{5pt}
\setlength{\abovecaptionskip}{2pt} 
\renewcommand{\arraystretch}{0.95}
\centering
\begin{tabular}{@{}l cccc c@{}}
\toprule
\multirow{2}{*}{\textbf{Method}} & \multicolumn{4}{c}{Enforcing flat at height $z$} & \multirow{2}{*}{\makecell{DSO \\ (Ours)}} \\
\cmidrule(lr){2-5}
& $z=0.05$ & $z=0.1$ & $z=0.15$ & $z=0.2$ & \\
\midrule
$\%$ Stable      & $94.2$ & $90.5$ & $93.2$ & $\underline{95.8}$ & $\mathbf{99.0}$ \\
Chamfer Distance & $\underline{0.0502}$ & $0.0537$ & $0.0591$ & $0.0662$ & $\mathbf{0.0440}$ \\
\bottomrule
\end{tabular}
\caption{
\textbf{Comparison} with post-processing baselines.
}
\label{tab:additional-comp}
\end{table}

\begin{figure}[tb!]
\centering
\includegraphics[width=\columnwidth, clip, trim=0 10 0 0]{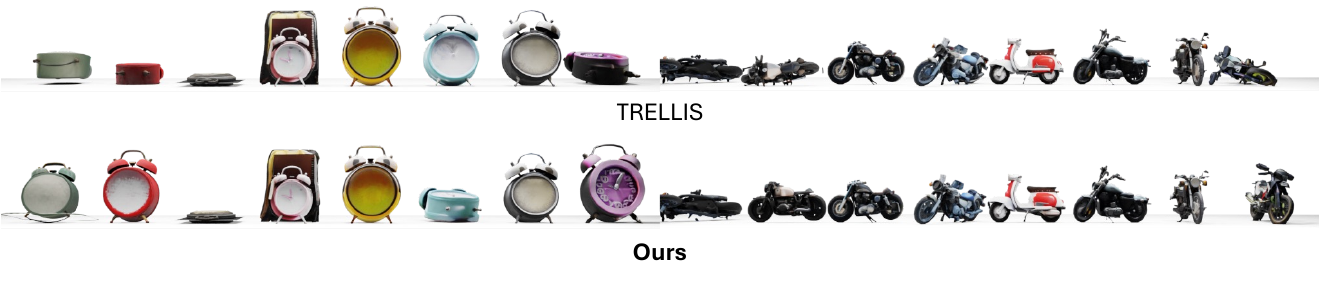}
\caption{
\method fine-tuned TRELLIS (\textbf{ours}) is more likely to generate physically sound 3D objects when conditioned on \emph{real-world} images of challenging categories.
}%
\label{fig:real-comparison}
\end{figure}
\subsection{Additional Results on In-the-Wild Images}
To assess the generalization of \method fine-tuned models in generating physically sound 3D objects from real-world images, we curate a set of $30$ CC-licensed images for each category: stable alarm clocks and motorcycles supported by kickstands. We select these two categories because the base model, TRELLIS, struggles to generate physically stable versions of these objects. The results are reported in~\cref{tab:compare-real}, with \emph{randomly sampled} examples visualized in \cref{fig:real-comparison}.
As is evident, \method enhances the model's ability to generate assets that remain stable under gravity from in-the-wild images of stable objects.

\section{Additional Discussions}

\begin{figure}[htbp]
\small
\setlength{\abovecaptionskip}{2pt} 
\centering
\includegraphics[width=\columnwidth]{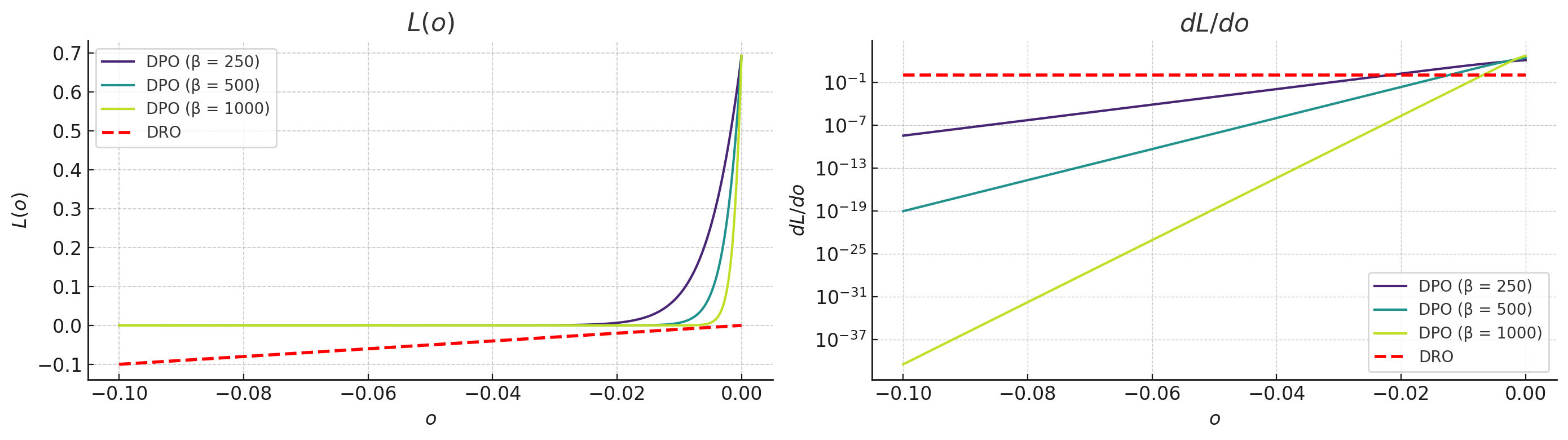}
\caption{\textbf{Plots} of $\mathcal{L}_\text{DRO}$ and $\mathcal{L}_\text{DPO}$ and their derivatives.
}%
\label{fig:analysis}
\end{figure}

\paragraph{A deeper analysis of DRO \vs DPO.}
We further analyze the similarities and differences between $\mathcal{L}_{\text{DRO}}$ and $\mathcal{L}_{\text{DPO}}$.
Both losses are monotonic functions of $o = \Vert \bm{\epsilon}^w - \bm{\epsilon}_\theta(\bm{x}_t^w, t)\Vert ^2 _2 - \Vert \bm{\epsilon}^w - \bm{\epsilon}_\text{ref}(\bm{x}_t^w, t) \Vert^2_2
-\left( \Vert \bm{\epsilon}^l - \bm{\epsilon}_\theta(\bm{x}_t^l, t) \Vert ^2_2 - \Vert \bm{\epsilon}^l - \bm{\epsilon}_\text{ref}(\bm{x}_t^l, t) \Vert ^2_2 \right)$.
In~\cref{fig:analysis}, we plot each loss (\textbf{left}) and its derivative with respect to $o$ (\textbf{right}, log-scale).
A key difference is that $\frac{d\mathcal{L}_\text{DRO}}{do}$ is constant, while $\frac{d\mathcal{L}_\text{DPO}}{do}$ decays exponentially as $o$ decreases.
As a result, $o$ tends to plateau during optimization of $\mathcal{L}_\text{DPO}$.
This leads to faster convergence with $\mathcal{L}_\text{DRO}$, although extended training may harm performance.

\paragraph{Scaling behaviors when optimizing $\mathcal{L}_{\text{DRO}}$.}
In~\cref{sec:exp_scaling}, we analyzed how DSO scales when optimizing $\mathcal{L}_\text{DPO}$.
Here, we present the corresponding scaling behavior for $\mathcal{L}_{\text{DRO}}$.
As shown in~\cref{tab:DRO-scaling}, performance peaks at $4,000$ training steps, after which the geometry quality noticeably degrades---consistent with our earlier analysis.
Scaling with training data follows a similar trend to that observed for $\mathcal{L}_\text{DPO}$ in~\cref{fig:scaling_data}.

\begin{table}[ht!]
\small
\tablestyle{3.0pt}{1.0}
\setlength\tabcolsep{5pt}
\setlength{\abovecaptionskip}{2pt} 
\renewcommand{\arraystretch}{1.0}
\begin{tabular}{@{}l cccc@{}}
\toprule
Training steps & $2000$ & $3000$ & $4000$ & $5000$ \\
\midrule
$\%$ Stable & $91.5$ & $96.9$ & $\mathbf{99.0}$ & $98.7$ \\
Chamfer D. & $0.0473$ & $0.0464$ & $\mathbf{0.0440}$ & $\textcolor{red}{0.0853}$ \\
\bottomrule
\end{tabular}
\caption{
\textbf{Scaling behavior with training compute} of $\mathcal{L}_{\text{DRO}}$.
}%
\label{tab:DRO-scaling}
\end{table}

\section{Limitations and Future Work}
DSO's self-improving scheme relies on the base model generating at least some positive samples, and hence may be less effective for base models where such samples are rare.
DSO opens up new possibilities for integrating physical constraints into generative models, enhancing their applicability in real-world scenarios where adherence to such constraints is crucial.

\end{document}